\documentclass[lettersize,journal]{IEEEtran}
\usepackage{amsmath,amsfonts,amsthm}
\usepackage{amssymb}
\usepackage{bm}
\usepackage{array}
\usepackage{textcomp}
\usepackage{stfloats}
\usepackage{url}
\usepackage{verbatim}
\usepackage{graphicx}
\usepackage{booktabs}
\usepackage{multirow}
\usepackage{multicol}
\usepackage{xcolor}
\usepackage{colortbl}
\usepackage{bbding}
\usepackage{tabularx}
\usepackage{pgfplots}
\usepackage{xcolor}
\usepackage[numbers,sort&compress]{natbib}
\usepackage{lettrine}
\usepackage[caption=false,font=footnotesize,labelfont=rm,textfont=rm]{subfig}

\usepackage{nicefrac}

\definecolor{cvprblue}{rgb}{0.21,0.49,0.74}
\usepackage[pagebackref,breaklinks,colorlinks,allcolors=cvprblue]{hyperref}
\usepackage[capitalize]{cleveref}

\begin{document}

\title{Explore Intrinsic Geometry for Query-based Tiny and Oriented Object Detector with Momentum-based Bipartite Matching}

\author{ Junpeng Zhang,~\IEEEmembership{Member, IEEE},
Zewei Yang,
Jie Feng$^{*}$,~\IEEEmembership{Senior Member, IEEE},
Yuhui Zheng,~\IEEEmembership{Member, IEEE},
Ronghua Shang,~\IEEEmembership{Senior Member, IEEE},
Mengxuan Zhang,~\IEEEmembership{Member, IEEE}}

\markboth{Journal of \LaTeX\ Class Files,~Vol.~14, No.~8, August~2021}%
{Shell \MakeLowercase{\textit{et al.}}: A Sample Article Using IEEEtran.cls for IEEE Journals}

\maketitle

\begin{abstract}
Recent query-based detectors have achieved remarkable progress, yet their performance remains constrained when handling objects with arbitrary orientations, especially for tiny objects capturing limited texture information.
This limitation primarily stems from the underutilization of intrinsic geometry during pixel-based feature decoding and the occurrence of inter-stage matching inconsistency caused by stage-wise bipartite matching. 
To tackle these challenges, we present IGOFormer, a novel query-based oriented object detector that explicitly integrates intrinsic geometry into feature decoding and enhances inter-stage matching stability.
Specifically, we design an Intrinsic Geometry-aware Decoder, which enhances the object-related features conditioned on an object query by
injecting complementary geometric embeddings extrapolated from their correlations to capture the geometric layout of the object, thereby offering a critical geometric insight into its orientation.
Meanwhile, a Momentum-based Bipartite Matching scheme is developed to adaptively aggregate historical matching costs by formulating an exponential moving average with query-specific smoothing factors, effectively preventing conflicting supervisory signals arising from inter-stage matching inconsistency.
Extensive experiments and ablation studies demonstrate the superiority of our IGOFormer for aerial oriented object detection, achieving an AP$_{50}$ score of 78.00\% on DOTA-V1.0 using Swin-T backbone under the single-scale setting. 
The code will be made publicly available.
\end{abstract}


\section{Introduction}

\begin{figure}[t]
\centering
\includegraphics[width=0.95\linewidth]{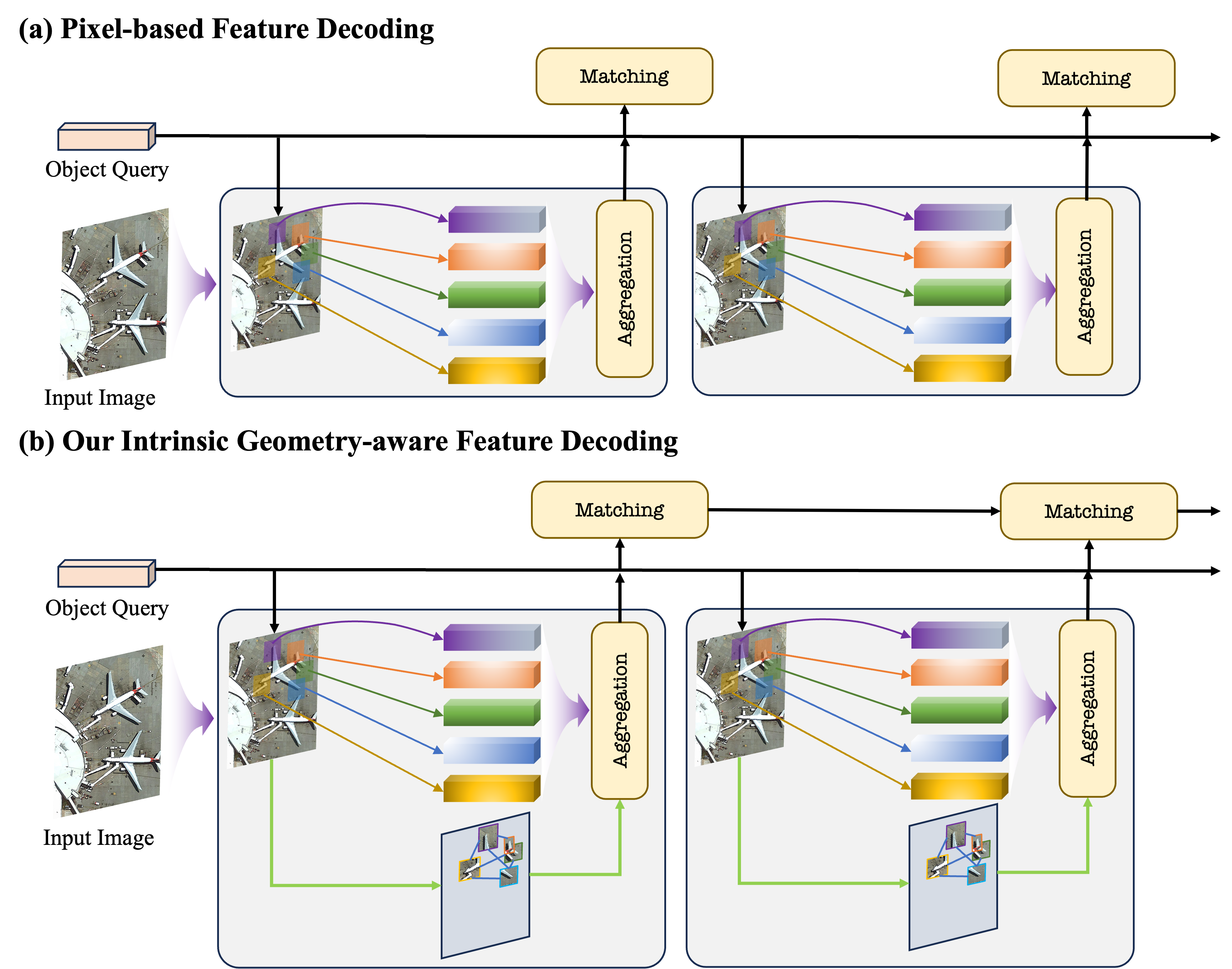}
\caption{Comparison against conventional pixel-based feature decoding in query-based detectors. (a), given an input image, an object query manages to identify a set of representative object-related features, which are then aggregated for query refinement. In (b), we argue that the relative relations among identified object-related features provide stronger evidence of the object's orientation and spatial arrangement. To capitalize on these relations, we propose an Intrinsic Geometry-aware decoding module, which explicitly explores these relationships to enhance the query's representativeness during iterative query refinement.}
\label{fig:comp_between_cross_attn}
\end{figure}

\IEEEPARstart{O}{riented} object detection enhances the capability of object detection by delineating each object in an image with a rotated bounding box, providing a more precise and comprehensive spatial description of the object.
This enriched geometric representation facilitates a wide range of applications where accurate object layouts are critical, such as surveillance systems~\cite{shi2018face,kaewkorn2026face} and earth observation~\cite{xia2018dota,fsd_tpami_2025}. 
Despite recent advancements in object detection~\cite{ren2015fasterrcnn, fcos_iccv2019, ssd_eccv2016,carion2020endtoenddetr}, detecting oriented objects remains a highly challenging task, due to the inconsistent feature alignment caused by variations in object orientations.
This challenge is further amplified in scenarios dominated by tiny objects, such as those commonly encountered in aerial images, where limited object scale results in insufficient appearance cues for both accurate recognition and precise localization.

Recently, the emergence of DEtection TRansformer (DETR) ~\cite{carion2020endtoenddetr} has facilitated the development of a new class of query-based detectors, achieving significant performance gains in generic object detection \cite{carion2020endtoenddetr, deformable_detr_iclr2020, liu2022dabdetr}.
The success of these query-based detectors primarily stems from their enhanced feature decoding capability as well as the progressive refinement mechanism.
However, in the context of tiny and oriented object detection, arbitrary object orientations and insufficient texture information of tiny objects pose challenges to decoding effectiveness, leading to unstable query refinement and suboptimal performance of query-based detectors.

\begin{figure*}[t]
\centering
\includegraphics[width=1.0\textwidth]{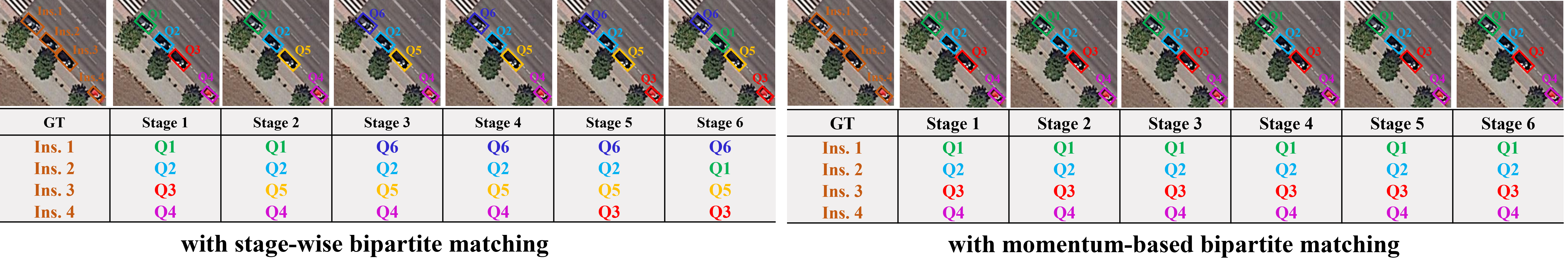}
\caption{An example on matching trajectories with stage-wise bipartite matching and our Momentum-based Bipartite Matching. Each color represents a different query, and the change of color for the same instance denotes the inter-stage identity shift across decoding stages.
}
\label{fig:mbm}
\end{figure*}

The attention-based multi-stage feature decoding scheme plays a fundamental role in query-based detectors, where both self-attention and cross-attention mechanisms are employed to model the global inter-query dependencies and aggregate progressive refinements to object queries at each stage.
Recent works in query-based object detectors have explored various cross-attention designs to enhance the decoding capabilities, where each object query attends to the image feature for identifying object-related features via a similarity map \cite{carion2020endtoenddetr} or from a sparse set of query-conditioned sampling locations \cite{deformable_detr_iclr2020, gao2022adamixer}.
Query refinement is then achieved by aggregating these object-related features, typically through weighted aggregation, as illustrated in \cref{fig:comp_between_cross_attn}.
However, these methods are primarily designed for axis-aligned objects in natural images and exhibit poor generalizability to objects with arbitrary orientations, due to their inefficiency in handling feature misalignment for oriented objects.
To mitigate this issue, orientation-aware sampling techniques are developed to align the sampling locations with object orientation \cite{arsdetr_cvpr2023, dai2022ao2detr, zhao2024orientedformer}. 
Despite these advancements, their performance remains constrained on tiny and oriented objects.
A key factor contributing to this limitation is that the limited texture information extracted from a few pixels turns insufficient in inferring accurate object orientations, whereas the geometric structures, offering crucial clues for accurate oriented object localization, are frequently overlooked.

Another critical factor limiting the performance of query-based object detectors for tiny and oriented objects is the conflicting supervisory signals that arise from the inter-stage matching inconsistency, due to the stage-wise bipartite matching.
Query-based detectors rely on a progressive query refinement process to gradually enhance the spatial alignment between object predictions and their corresponding groundtruths. 
However, due to the inherent instability in the predictions during training, an object is often matched with different queries at different decoding stages, resulting in ambiguous and inconsistent optimization objectives.
This identity shift issue is further exacerbated in tiny and oriented object detection, where arbitrary orientations and limited texture information significantly increase prediction instability, causing frequent alternations in object-query assignments across decoding stages. 
As visualized in \cref{fig:mbm}, the object queries $\mathbf{Q}_1$ and $\mathbf{Q}_3$ initially matched with $\text{Inst.1}$ and $\text{Inst.3}$ at the 1st decoding stage are matched with $\text{Inst.2}$ and $\text{Inst.4}$ at the final stage, whose initial locations are distant from the final predictions.
To mitigate the ambiguities arising from inconsistent label assignments across decoding stages, recent works have explored surrogate solutions that accelerate the network convergence by reducing such inconsistency, typically through the introduction of auxiliary denoising stages \cite{zhang2022dino, liu2023stable-dino,chen2023group,jia2023detrs}.
Despite their progress, these approaches overlook the fact that stable optimization fundamentally depends on consistent target assignments across stages, hindering their effectiveness by the lack of inter-stage communication in their stage-wise bipartite matching.

\begin{figure*}[t]
\centering
\includegraphics[width=1.0\textwidth]{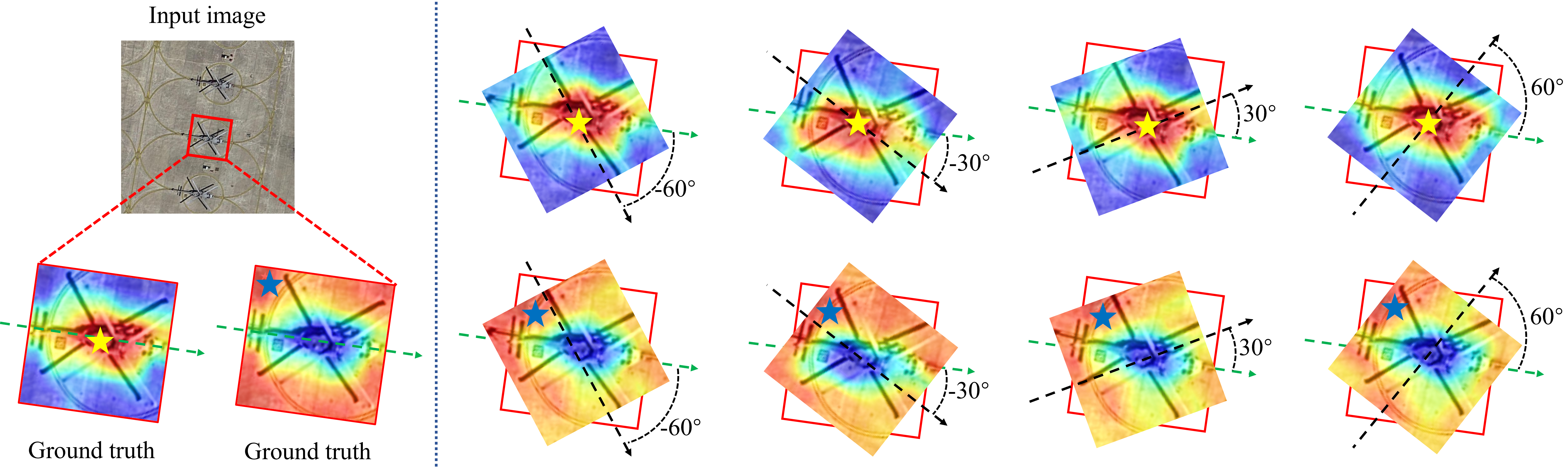}
\caption{Visualization of the correlation among object-related features of a helicopter. 
Two anchor features are marked by $\color{yellow}\bigstar$ and $\color{blue}\bigstar$, respectively.
Their correlations with other object-related features for the same object query are visualized in the first and second rows, respectively. 
As the bounding box is rotated to reflect the misalignment with the ground-truth bounding box, the highlighted regions persistently reveal the spatial and structural alignment of the object, while their distribution within the box remains sensitive to orientation changes.
}
\label{fig:relation_visualize}
\end{figure*}

To address the aforementioned issues, we propose IGOFormer, a novel query-based oriented object detector for tiny and oriented objects by exploring the objects' geometric structure in query refinement and enhancing inter-stage communication in bipartite matching. 
In this work, we highlight that an object's intrinsic geometry offers a crucial insight for object orientation, especially for objects with complex structures, such as helicopters.
As visualized in \cref{fig:relation_visualize}, the correlations among the object-related features identified by an input query reveal the spatial and structural alignment of the object, while remaining sensitive to its orientation, even when a tiny object captures a few pixels and exhibits limited texture information. 
Notably, as an object undergoes varying rotations, the feature correlations maintain strong alignment with the object's orientation, providing an important cue for correcting the predicted rotated bounding box.
To this end, we propose an Intrinsic Geometry-aware Decoder, where the object-related features identified by an input query are enriched with complementary geometric embeddings extracted from their correlations, as illustrated in \cref{fig:comp_between_cross_attn}.
For enhancing the interaction between object query and the encoded geometric embeddings, we utilize a dynamic neural network for generating query-conditioned filters for suppressing irrelevant information, where a Multi-group Scheme is designed for facilitating the filter diversity and reducing computational cost at the same time.
Meanwhile, instead of conducting bipartite matching at each decoding stage independently, we design a Momentum-based Bipartite Matching scheme that adaptively incorporates current matching cost with those from preceding stages.
This matching formulation naturally represents an exponential moving average of the matching cost throughout the decoder stages, thus enhancing inter-stage matching consistency and suppressing ambiguous optimization objectives during training.

The main contributions of this article are summarized as follows:
\begin{enumerate}
\item We introduce IGOFormer that leverages intrinsic geometry and inter-stage communication in query-based detector for precise tiny and oriented object detection.
\item Our Intrinsic Geometry-aware Decoder integrates an Intrinsic Geometry Augmentation process that approaches query refinement from a geometric perspective and encodes object's geometric structure by exploring the correlation among object-related features, thus providing a critical complementary cue for tiny and oriented object detection.
\item Our Momentum-based Bipartite Matching scheme adaptively aggregates historical matching costs and constructs an exponential moving average mechanism for suppressing drastic label assignment changes, effectively preventing conflicting supervisory signals arising from inter-stage matching inconsistency.
\end{enumerate}

To comprehensively validate the effectiveness of our IGOFormer for tiny and oriented object detection, we conduct extensive experiments on multiple challenging benchmarks, including DOTA-V1.0, DOTA-V1.5~\cite{xia2018dota}, and DIOR-R~\cite{aopg_tgrs2022}. 
Specifically, our method achieves an AP$_{50}$ score of 78.00\% on DOTA-V1.0 using Swin-T backbone under the single-scale setting.

\section{Related Work}

\subsection{Query-based Object Detector}

With recent advancements in attention mechanism, the emergence of DEtection TRansformer (DETR)~\cite{carion2020endtoenddetr} has facilitated the development of a new class of query-based detectors, achieving significant performance gains in generic object detection~\cite{carion2020endtoenddetr, deformable_detr_iclr2020, liu2022dabdetr}.
At its core, DETR employs a transformer-based architecture, integrating a multi-stage decoder for progressively refining a set of object queries through both self-attention and cross-attention mechanisms.
Built upon this design, Anchor-DETR~\cite{wang2022anchordetr} incorporates anchor boxes into object queries and offers predefined spatial locations to densify the query space, thereby ensuring improved initialization of object queries.
For further enhancing the adaptivity of object queries to randomness of objects' spatial arrangement in an image, Conditional-DETR~\cite{meng2021conditionaldetr} adopts an adaptive query initialization scheme that conditions object queries on image content, enabling the query-based detector to handle variations in object layout with large-scale and complex scenes.

In addition to these advancements in query design, cross-attention is another critical component in query-based detectors, where refinements to object queries are adaptively aggregated through the interaction between object queries and image features.
The vanilla cross-attention attends each object query to image features for identifying object-related features via a global similarity map, then aggregates query refinement via a weighted sum over all spatial locations in the feature map~\cite{carion2020endtoenddetr}.
To handle the anisotropy inherent in objects, DAB-DETR~\cite{liu2022dabdetr} introduces aspect-aware modulation factors into the positional encoding of cross-attention, leading to improved feature alignment for objects with varying aspect ratios.
To alleviate the computational burden of constructing a global similarity map for each query, Deformable DETR~\cite{deformable_detr_iclr2020} designs a computationally efficient solution that estimates a sparse set of sampling locations conditioned on each query, from which query refinement is collected from a multi-scale image feature pyramid.
For further enhancing the interaction between object query and image features, dynamic neural networks are incorporated to generate query-conditioned filters, thus leading to improved decoding capabilities~\cite{sun2021sparsercnn,gao2022adamixer}.
However, these detectors are primarily designed for axis-aligned objects in natural images and exhibit poor generalizability to oriented object detection, due to their inefficiency in handling feature misalignment for objects with arbitrary orientations.

\subsection{Oriented Object Detection}

Oriented object detection is a highly featured task in computer vision that aims to localize objects with arbitrary orientations and assign their category labels in an image. 
Significant efforts have been dedicated to this field, especially in the context of surveillance systems and Earth observation~\cite{xia2018dota, fair1m_isprs_2022}, where detection performance is challenged by arbitrary object orientations, small object scales, dense spatial distributions and complex backgrounds.
Built on the success of generic object detection~\cite{ren2015fasterrcnn,fcos_iccv2019}, the majority of oriented object detectors commonly follow a candidate-based paradigm, where redundant object candidates are generated via dense points~\cite{s2anet_tgrs2021, bbavector_wacv2021} or anchor priors~\cite{roi_trans_cvpr2019, xie2021orientedrcnn}, and utilize an additional regression head to minimize spatial misalignment between the predicted rotated bounding boxes and their assigned ground-truth.

To address the boundary discontinuity issue in orientation estimation, two main types of localization losses have been developed: the classification-based loss, which reformulates orientation regression into a classification problem~\cite{cls_eccv2020,dcl_cvpr2021}, and the distribution-based loss, which parameterizes each rotated bounding box as a probability distribution~\cite{yang2021rethinkingrotatedgwd,KLD,kfiou_iclr2022}.

Alternatively, some studies attempt to model oriented object localization as a convex-hull search problem, thus each rotated bounding box can be inferred from a set of representative points \cite{cfa_cvpr2021, li2022orientedreppoints, point_axes_eccv2024}. 
Despite the performance gains brought by these improved network architectures and loss designs, these methods still heavily rely on hand-crafted priors and heuristic post-processing such as non-maximum suppression for removing redundant predictions, which limits end-to-end learning of the entire detection pipeline.

Inspired by the remarkable progress in recent query-based detectors, the effective attention mechanism is adopted for constructing performant oriented object detectors, enabling fully end-to-end network learning by leveraging the set prediction paradigm~\cite{dai2022ao2detr, arsdetr_cvpr2023, point_axes_eccv2024}.
To enhance the feature alignment on oriented objects, orientation-aware cross-attention module is developed by incorporating object orientation in identifying object-related features~\cite{arsdetr_cvpr2023,zhang2024orienteddino}.
OrientedFormer~\cite{zhao2024orientedformer} transforms each predicted bounding box to a rotated gaussian kernel as an auxiliary guidance to conventional positional encoding, thus enhancing the query representation by improved feature alignment.
Oriented-DETR~\cite{point_axes_eccv2024} develops a location and rotation decoupling scheme, where interactions between identified object-related features are enabled by a group self-attention mechanism. 
Despite these advances, their performance on tiny and oriented objects remains constrained, due to the limited texture information available.
In this work, we target incorporating the object's geometric information to enrich object-related features for tiny and oriented object detection.

\subsection{Inter-stage Inconsistency in Bipartite Matching}

Bipartite matching plays a central role in end-to-end learning for query-based object detectors. 
In these models, bipartite matching establishes a one-to-one assignment between object queries and ground-truth, thus allowing the model to directly optimize detection results by eliminating the need for non-differentiable non-maximum suppression.
However, the matching inconsistency by bipartite matching across decoding stages has been identified as a critical bottleneck limiting the performance of query-based detectors~\cite{liu2023stable-dino,li2022dndetr,zhang2022dino}. 
Existing works commonly attribute such instability to the slow convergence.
To mitigate this issue, a widely adopted solution is to employ one or more auxiliary decoders, where extra supervisory signals by both one-to-one and one-to-many assignments are imposed to facilitate the network convergence~\cite{zhang2022dino, liu2023stable-dino,hu2023dacdetr,zhao2024msdetr}.
This inter-stage inconsistency issue turns particularly pronounced in tiny and oriented object detection, as the arbitrary object orientations and small object scales make stable predictions even more challenging. 
On aerial images, EMO2-DETR~\cite{hu2023emo2detr} attempts to mitigate this issue by focusing on high-quality negative samples in bipartite matching. 
In this work, we manage to address identity shifts in stage-wise bipartite matching from a different perspective \textemdash improving matching stability by maintaining smooth matching costs and progressively updating them across decoding stages.

\section{Method}

\begin{figure*}[t]
\centering
\includegraphics[width=.985\textwidth]{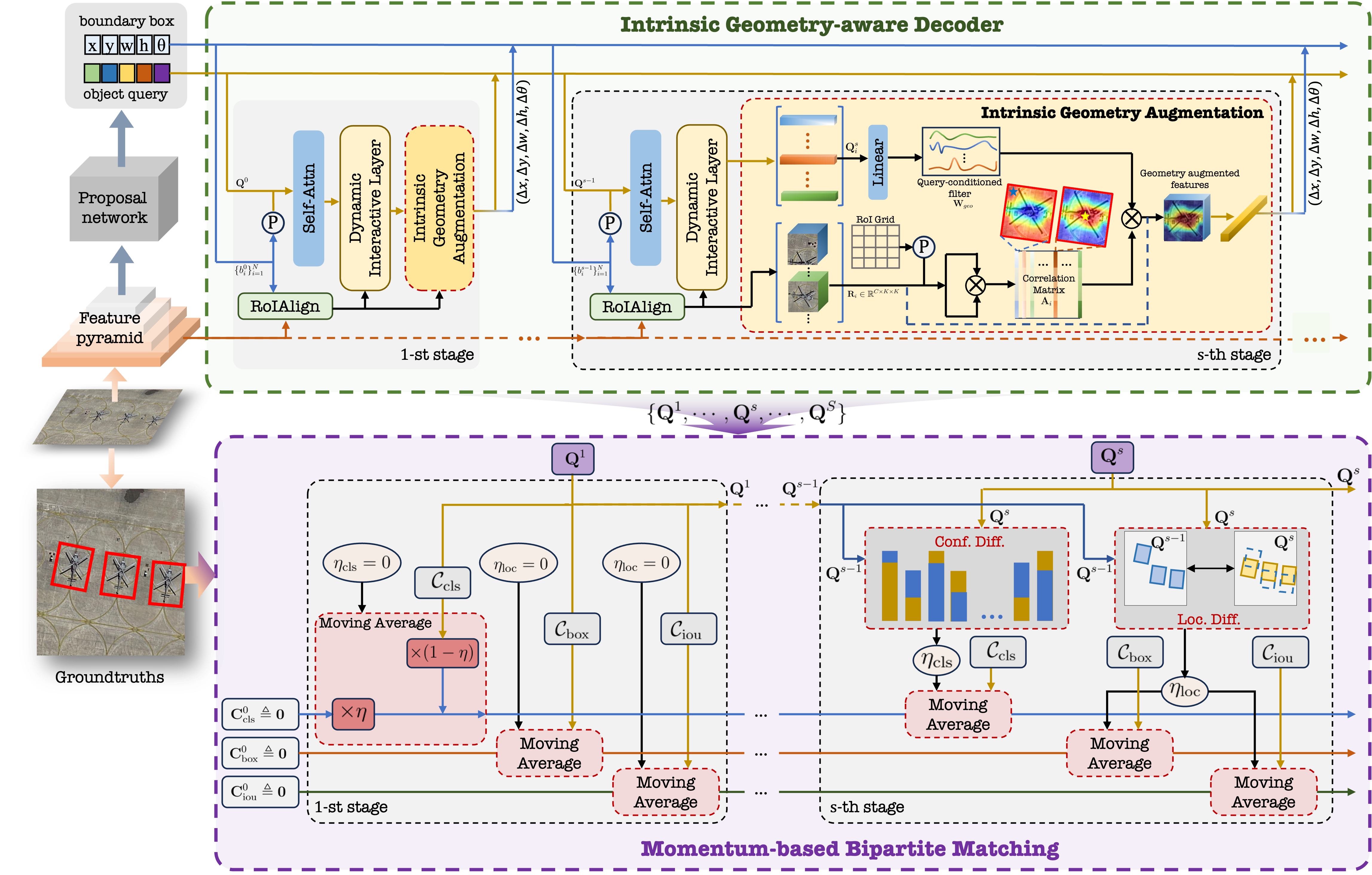} 
\caption{Model architecture of the proposed IGOFormer. Our IGOFormer follows the encoder-free architecture in \cite{sun2021sparsercnn}, and an Intrinsic Geometry Augmentation module and a Momentum-based Bipartite Matching scheme are introduced. The Intrinsic Geometry Augmentation explores the correlation among object-related features, thus providing a critical complementary cue for tiny and oriented object detection. Meanwhile, our Momentum-based Bipartite Matching scheme adaptively aggregates historical matching costs and constructs an exponential moving average mechanism for suppressing drastic label assignment changes.}
\label{fig:model_arch}
\end{figure*}

\subsection{Query-based Detector Revisited}

Query-based detectors, originating from the seminal DETR framework~\cite{carion2020endtoenddetr}, have emerged as a dominant paradigm in modern object detection.
They typically employ an architecture comprising a backbone network, an optional image encoder, a multi-stage decoder, and prediction heads.
The recent success of these detectors hinges on two core principles: an iterative query refinement procedure and a bipartite matching scheme.

\textbf{Iterative Query Refinement.}
Given an input image $\mathbf{I}$, let $\bm{\mathcal{F}} \triangleq \{\mathbf{F}^\ell\}_{\ell=1}^L$ denote the $L$-level image feature pyramid extracted by the backbone and optional encoder, where $\mathbf{F}^\ell \in \mathbb{R}^{H_\ell \times W_\ell \times C_\ell}$ is the $\ell$-th feature map with spatial resolution $(H_\ell, W_\ell)$ and $C_\ell$ channels. 
The iterative feature decoding process progressively refines an initial set of 
$N$ object queries, $\mathbf{Q}^{0}$, across $S$ decoding stages.
These initialized queries $\mathbf{Q}^{0}$ are either instantiated as learnable embeddings~\cite{carion2020endtoenddetr} or generated dynamically via a lightweight proposal network~\cite{deformable_detr_iclr2020}.
Then the initialized queries are fed to a sequence of decoding stages for progressive refinement,
\begin{equation}
\mathbf{Q}^{s} = \text{Decoder}_s(\mathbf{Q}^{s-1} | \bm{\mathcal{F}}), \quad s \in \{1, \cdots, S\}.
\end{equation}

At each decoding stage $s$, the input object queries $\mathbf{Q}^{s-1}$ are updated by two attention modules.
A self-attention module is adopted for capturing the inter-query contextual dependencies, then a cross-attention module enables the queries to interact with the feature pyramid $\bm{\mathcal{F}}$, allowing them to incorporate semantically relevant features.
This refinement process is formulated as
\begin{equation}
\begin{aligned}
\mathbf{\widetilde{Q}}^{s} &= \text{Self-Attn}(\mathbf{Q}^{s-1}), \\
\mathbf{Q}^{s} &= \Gamma\left( \text{Cross-Attn}(\mathbf{\widetilde{Q}}^{s}, \bm{\mathcal{F}}) \right),
\end{aligned}
\end{equation}
where $\Gamma$ is a feed-forward network for projecting the aggregated features for the subsequent stage.
The iterative refinement process across $S$ stages allows the queries to evolve progressively, with each stage improving the query representations and facilitating better detection performance.

Within this process, cross-attention plays a crucial role in determining both \emph{which} image features are selected and \emph{how} they are incorporated into refined queries.
To this end, a variety of cross-attention designs have been explored for identifying object-related features, including similarity-based correlations~\cite{carion2020endtoenddetr}, query-conditioned sampling~\cite{deformable_detr_iclr2020}, or feature pooling from sparse proposals~\cite{sun2021sparsercnn}. 
These identified features are then aggregated via weighted sum~\cite{carion2020endtoenddetr,deformable_detr_iclr2020} or through additional dynamic networks conditioned on the input queries~\cite{sun2021sparsercnn}. 
Despite their advances, existing designs largely emphasize discovering and combining semantic cues in a pointwise manner, while paying limited attention to encoding the geometric structure of objects, which is critical for objects with arbitrary orientation.


\textbf{Bipartite Matching for End-to-end Training.}
To construct a fully end-to-end training procedure, query-based detectors enforce one-to-one assignment between predicted queries and ground-truth, formulating a bipartite matching problem. 
Typically, bipartite matching is performed at each decoding stage independently, by solving a minimum-cost assignment between that stage’s queries and the ground-truth.

Let $\mathcal{G}=\{(\hat{y}_j,\hat{b}_j)\}_{j=1}^{|\mathcal{G}|}$ denote the ground-truth set for an image $\mathbf{I}$, comprising pairs of object categories $\hat{y}_j$ and its corresponding rotated bounding boxes $\hat{b}_j$.
The corresponding prediction set from $\mathbf{Q}^s$ at a decoding stage $s$ is denoted by $\{(p_i^s,b_i^s)\}_{i=1}^{N}$, where $p_i^s$ and $b_i^s$ represent the predicted classification score and bounding box, respectively.
A pairwise cost matrix $\mathbf{C}^s \in \mathbb{R}^{N\times|\mathcal{G}|}$ is constructed, where the entry $\mathbf{C}^s_{ij}$ measures how well the $i$-th prediction matches $j$-th ground-truth,
\begin{equation}
\mathbf{C}^s_{ij}
= \lambda_{\mathrm{cls}}\, \mathcal{C}_{\mathrm{cls}}(p_i^s,\hat{y}_j)
+ \lambda_{\mathrm{box}}\, \mathcal{C}_{\mathrm{box}}(b_i^s,\hat{b}_j)
+ \lambda_{\mathrm{iou}}\, \mathcal{C}_{\mathrm{iou}}(b_i^s,\hat{b}_j),
\end{equation}
where $\mathcal{C}_{\mathrm{cls}}$, $\mathcal{C}_{\mathrm{box}}$ and $\mathcal{C}_{\mathrm{iou}}$ represent the classification, bounding box, and IoU cost terms, respectively, and $\lambda_{\mathrm{cls}}$, $\lambda_{\mathrm{box}}$ and $\lambda_{\mathrm{iou}}$ are their corresponding weight coefficients.

The bipartite matching at stage $s$ is then obtained by solving a one-to-one assignment that minimizes the total matching cost,
\begin{equation}
\mathcal{A}^s = \arg\min_{\mathcal{A}}\sum_{j=1}^{|\mathcal{G}|}\mathbf{C}^s_{\mathcal{A}(j),j},
\end{equation}
where $\mathcal{A}$ is an injective mapping from ground-truth to predictions, and its optimal solution is commonly obtained using the Hungarian algorithm.

While the stage-wise matching allows flexibility in reassessing assignments at each decoding stage, it also introduces a systemic instability in multi-stage decoding pipelines, \textit{i.e.}, the identity shift issue, which stems from the prediction fluctuations across stages. 
This issue becomes more pronounced with oriented objects, where the arbitrary object orientations impose stronger requirements on matching consistency across stages to stabilize the training process.

\textbf{Summary.}
For oriented object detection, a performant query-based detector requires enhanced capability to encode the geometric structure of objects during decoding, along with improved inter-stage matching consistency during training.
Both requirements motivate our design of IGOFormer presented in the following subsections.

\subsection{Overview of IGOFormer}

To address limited appearance cues and identity shifts inherent in oriented object detection, we introduce IGOFormer, a novel query-based oriented object detector, by leveraging the geometric structures of objects and enforcing inter-stage matching consistency.
To this end, we propose an Intrinsic Geometry-aware Decoder, where an Intrinsic Geometry Augmentation process explicitly integrates geometric cues into each decoding stage.
At the same time, a Momentum-based Bipartite Matching scheme is developed by adaptively incorporating the matching cost from preceding stages, thus ensuring inter-stage matching consistency during training.

As illustrated in \cref{fig:model_arch}, our IGOFormer adopts an encoder-free architecture~\cite{sun2021sparsercnn, gao2022adamixer}, consisting of a backbone, a multi-stage decoder and a set of prediction heads. 
Given an input image $\mathbf{I}$, the backbone extracts a feature pyramid $\bm{\mathcal{F}}$.
The obtained feature pyramid $\bm{\mathcal{F}}$ is then fed to a lightweight proposal network to generate initial object queries. 
In this work, the proposal network contains a single $1\times 1$ convolutional layer for classification and a stack of three $1\times 1$ convolutional layers for box regression.
We select the Top-$N$ predictions per image from this network module for constructing the object queries $\mathbf{Q}^{0}$.

Next, the obtained initial object queries $\mathbf{Q}^{0}$, together with the feature pyramid $\bm{\mathcal{F}}$, are passed to our Intrinsic Geometry-aware Decoder for progressive refinement over $S$ stages.
At each stage, instead of solely relying on the semantic cues, our Intrinsic Geometry Augmentation explicitly injects intrinsic geometric information into the decoding process, enabling query refinement with both semantic and geometric guidance.
Finally, the refined queries $\mathbf{Q}^{S}$ are fed to the prediction heads to yield the detection outputs.

For mitigating the identity shift issue, the proposed Momentum-based Bipartite Matching scheme is introduced during training.
After each decoding stage, our scheme replaces stage-wise instantaneous matching with a momentum-style cost that integrates matching costs from prior stages, and passes this cost together with the current matching status to subsequent stages.
The remainder of this section details the proposed Intrinsic Geometry-aware Decoder in \cref{subsec:method_decoder} and our novel Momentum-based Bipartite Matching scheme in \cref{subsec:method_matching}.

\subsection{Intrinsic Geometry-aware Decoder}
\label{subsec:method_decoder}


To promote the inherent spatial and structural properties of an oriented object in query decoding, our Intrinsic Geometry-aware decoding module explores the correlations among object-related features to complement the semantic interaction between an input query and image features. 

As illustrated in \cref{fig:model_arch}, our decoder consists of a sequence of $S$ decoding stages, each of which employs both self-attention and cross-attention mechanisms for progressive query refinement.
Specifically, each decoding stage $s$ takes the refined object queries $\mathbf{Q}^{s-1}$ together with their corresponding predicted rotated bounding boxes $\{b_{i}^{s-1}\}_{i=1}^{N}$ as input.
Following the widely adopted decoding schemes~\cite{deformable_detr_iclr2020, carion2020endtoenddetr}, a self-attention module is first employed for capturing the inter-query dependency.
Simultaneously, for each updated object query $\mathbf{\widetilde{Q}}_i^{s}$, the object-related features inside its previously predicted bounding box $b_{i}^{s-1}$ are pooled from the feature pyramid $\bm{\mathcal{F}}$ using a Rotated RoIAlign operator, $\mathbf{R}_i^{s} = \mathrm{RoIAlign}(\bm{\mathcal{F}}, b_{i}^{s-1})$.
Then, the updated object queries $\mathbf{\widetilde{Q}}^{s}$, along with their corresponding object-related features $\bm{\mathcal{R}}^{s} = \{\mathbf{R}_i^{s}\}_{i=1}^{N}$, are passed to a dynamic interactive layer~\cite{sun2021sparsercnn}, where dynamic filters conditioned on each input query $\mathbf{\widetilde{Q}}_i^{s}$ are imposed for obtaining semantic refinements.
Finally, the proposed Intrinsic Geometry Augmentation is applied, which plays an important role in revealing the intrinsic geometry cues inside the object-related features and injecting them to construct more representative queries.
The decoding process at a single stage $s$ is formulated as
\begin{equation}
\begin{aligned}
\mathbf{\widetilde{Q}}^{s} &= \text{Self-Attn}(\mathbf{Q}^{s-1}), \\
\mathbf{Q}^{s} &= \mathcal{D}_{\mathrm{DI}}(\mathbf{\widetilde{Q}}^{s}, \bm{\mathcal{R}}^{s}), \\
\mathbf{Q}^{s} &= \Gamma\left( \mathcal{D}_{\mathrm{IGA}}(\mathbf{Q}^{s}, \bm{\mathcal{R}}^{s}) \right),
\end{aligned}
\end{equation}
where $\mathcal{D}_{\mathrm{DI}}(\cdot)$ and $\mathcal{D}_{\mathrm{IGA}}(\cdot)$ denote the dynamic interactive layer and our Intrinsic Geometry Augmentation respectively.

\textbf{Intrinsic Geometry Augmentation.} 
Inspired by the observation that the correlations among object-related features reveal the intrinsic geometry of an oriented object, our Intrinsic Geometry Augmentation enriches these features by encoding the correlations into complementary geometric embeddings, which are subsequently interacted with the input query.

For each input query $\mathbf{Q}_i^{s-1}$, its corresponding object-related features $\mathbf{R}_i^{s} \in \mathbb{R}^{C \times K \times K}$ are extracted from the image feature pyramid $\bm{\mathcal{F}}$ via RoIAlign, where $K$ is the spatial resolution of the pooled feature map.
First, we augment the object-related features in $\mathbf{R}_i$ with auxiliary positional information, $\bar{\mathbf{R}}_i = \mathbf{R}_i + \mathbf{P}_i$, where $\mathbf{P}_i \in \mathbb{R}^{C \times K \times K}$ represents the positional embeddings extracted using the sinusoidal encoding scheme~\cite{carion2020endtoenddetr}, mapping each normalized pixel coordinate to a $C$-dimensional embedding.
For simplicity, we omit the stage index $s$ in intermediate variables hereafter.

Then, the pairwise correlations among object-related features in $\bar{\mathbf{R}}_i$ are quantified by measuring their similarities,
\begin{equation}
\begin{aligned}
\bar{\mathbf{R}}_i =& \mathrm{Reshape}_{(C, K, K) \rightarrow (C, K \times K)}(\bar{\mathbf{R}}_i), \\
&\mathbf{A}_i = \sigma(\frac{ (\mathbf{W}_a \bar{\mathbf{R}}_i)^{T} \mathbf{W}_b \bar{\mathbf{R}}_i}{\sqrt{C}}),
\end{aligned}
\end{equation}
where $\mathbf{W}_a \in \mathbb{R}^{C \times C}$ and $\mathbf{W}_b \in \mathbb{R}^{C \times C}$ denote two independent linear projections, and $\sigma(\cdot)$ is the Softmax operator for normalizing the similarity score.
In the resulting correlation matrix $\mathbf{A}_i$, each column corresponds to the correlations between a certain feature and all others in $\bar{\mathbf{R}}_i$, effectively capturing the geometric structure of the object. 
Next, this correlation matrix $\mathbf{A}_i$ is projected to a $C$-dimensional geometric embedding via another linear layer, 
\begin{equation}
\mathbf{G}_i = \mathrm{Linear}(\mathbf{A}_i) \in \mathbb{R}^{C \times K^2}.
\end{equation}

Subsequently, we apply a filter conditioned on the input query $\mathbf{Q}_i^{s-1}$ for enhancing its interaction with the geometric embeddings and suppressing irrelevant features,
\begin{equation}
\begin{aligned}
\mathbf{W}_{geo} =& \mathrm{Reshape}_{C \times C \rightarrow (C, C)} \left( \text{Linear}(\mathbf{Q}_i^{s-1}) \right), \\
\hat{\mathbf{G}}_i =& \mathbf{W}_{geo} \mathbf{G}_i.
\end{aligned}
\end{equation}
This set of geometric embeddings $\hat{\mathbf{G}}_i$ captures the relative geometric relation among different parts of an object, which gives crucial cues for enhancing the representation of oriented objects.

Finally, these geometric embeddings are flattened and projected to construct a refined object query,
\begin{equation}
\begin{aligned}
& \hat{\mathbf{G}}_i = \mathrm{Reshape}_{(C, K, K) \rightarrow C \times K \times K}(\mathbf{R}_i + \hat{\mathbf{G}}_i), \\
& \mathbf{Q}_i^s = \mathrm{Linear}(\hat{\mathbf{G}}_i).
\end{aligned}
\end{equation}
This geometric feature enhancement mechanism enables the model to adaptively reinforce object geometric structures based on the query's semantic context, while preserving original feature information through residual connections from the original object-related features, thereby establishing mutually complementary representations that synergize semantic and geometric information.



\textbf{Multi-group Scheme.} 
Instead of relying on a single query-conditioned filter, we introduce a multiple-group scheme into our decoder, where the feature space is partitioned into multiple groups and a set of filters are generated accordingly for enhancing the diversity during query and feature interaction.

For the proposed Intrinsic geometric augmentation, the input queries $\mathbf{Q}^{s-1}$ and their corresponding object-related features $\bm{\mathcal{R}}^{s}$ are split into $M$ groups along the channel dimension, with each group occupying $d = \frac{C}{M}$ exclusive channels.
For each group $m \in \{1, \cdots, M\}$, a pair of query split $\mathbf{Q}_{m}^{s}$ and object-related feature split $\bm{\mathcal{R}}_{m}^{s}$ is defined as
\begin{equation}
\begin{cases}
\mathbf{Q}_{m}^{s} \triangleq \mathbf{Q}^{s}[:, d \times (m-1) : d \times m] \\
\bm{\mathcal{R}}_{m}^{s} \triangleq \left\{\mathbf{R}_{i}^{s}[d \times (m-1) : d \times m]\right\}_{i=1}^{N}
\end{cases}.
\end{equation}
Then, each pair of $\mathbf{Q}_{m}^{s}$ and $\bm{\mathcal{R}}_{m}^{s}$ is passed through an independent intrinsic geometric augmentation process,
\begin{equation}
\mathbf{Q}_{m}^{s} = \mathcal{D}_{\mathrm{IGA}}(\mathbf{Q}_{m}^{s}, \bm{\mathcal{R}}_{m}^{s}), \quad \forall m \in \{1, \cdots, M\}. \\
\end{equation}
Finally, we collect refined object queries from all groups, and stack them for reconstructing final output, 
\begin{equation}
\mathbf{Q}^{s} = \mathrm{Concat}(\mathbf{Q}_{1}^{s}, \cdots, \mathbf{Q}_{m}^{s}, \cdots, \mathbf{Q}_{M}^{s}),
\end{equation}
where $\mathrm{Concat}(\cdots)$ denotes concatenation operation along the channel dimension.

Similarly, we also adopt the Multi-group Scheme to the dynamic interactive layer, constructing a Multi-group Dynamic Interactive layer (MDI).
This Multi-group Scheme enables our decoder to capture more diverse information while reducing computational overhead, enabling differentiated and efficient refinement of oriented object queries.

\subsection{Momentum-based Bipartite Matching}
\label{subsec:method_matching}

The vanilla bipartite matching exhibits significant matching instability across decoding stages, due to its oversight of the matching status from preceding stages. 
Such instability risks generating contradictory supervision signals during optimization, ultimately degrading the detection performance.
To address this issue, we propose a Momentum-based Bipartite Matching (MBM) strategy by maintaining smooth and progressively updated matching costs across decoding stages.
The key idea is to incorporate the matching cost from the preceding stages while giving sufficient emphasis to the current matching costs.

Specifically, we define a momentum-based matching cost at the $s$-th decoding stage as a weighted combination of the matching costs at current stage as well as its preceding stage,
\begin{equation}
\label{eq:ema_cost}
\hat{\mathbf{C}}^s = \eta \hat{\mathbf{C}}^{s-1} + (1 - \eta) \mathbf{C}^{s}, \quad s \in \{1, \cdots, S\},
\end{equation}
where $\eta$ is the smoothing factor, and $\mathbf{C}^{0}$ is initialized as a zero matrix.
This formulation naturally represents an exponential moving average of the matching cost throughout the decoding stages, diminishing drastic cost changes due to prediction outliers.

Instead of using a consistent smoothing factor, we introduce adaptive smoothing factors for the classification costs and localization costs, tailored for each query independently. 
For the $i$-th query, let $p_i^s \in \mathbb{R}^{K+1}$ and $b_i^s \in \mathbb{R}^{5}$ denote its predicted classification scores and rotated bounding boxes at $s$-th decoding stage, where $K$ is the number of classes.
The adaptive smoothing factor for the classification cost is defined as
\begin{equation}
\eta_{cls}^{i, s} = 1 - e^{-\alpha \mathbb{D}_{KL}(p_i^s \parallel p_i^{s-1})},
\end{equation}
where $\mathbb{D}_{\mathrm{KL}}(p_i^s \parallel p_i^{s-1})$ denotes the Kullback-Leibler divergence between the probability distributions $p_i^s$ and $p_i^{s-1}$, and $\alpha > 0$ is a hyperparameter that controls the sensitivity of the smoothing factor to distribution changes.
When the normalized classification scores at $i$-th stage differ significantly from those at the immediately preceding stage, which is more likely caused by inherent instability in the prediction, a larger $\eta_{cls}^{i,s}$ is assigned to put more emphasis on its historical matching states, thus preventing conflicting supervisory signals in progressive query refinement.

Similarly, a query-specified adaptive smoothing factor is defined for the localization costs $\hat{\mathbf{C}}_{box}^s$ and $\hat{\mathbf{C}}_{iou}^s$,
\begin{equation}
\eta_{loc}^{i,s} = 1 - \text{IoU}(b_i^s, b_i^{s-1}),
\end{equation}
where $\text{IoU}(b_i^s, b_i^{s-1})$ measures the overlap between the bounding boxes $b_i^s$ and $b_i^{s-1}$. 
For the predicted bounding boxes with minimal inter-stage offsets, a smaller $\eta_{loc}^{i,s}$ helps assign more emphasis to the localization costs at the current stage.

By substituting the consistent smoothing factor $\eta$ in \cref{eq:ema_cost} with these query-specific adaptive factors $\eta_{cls}^{i, s}$ and $\eta_{loc}^{i, s}$, the overall Momentum-based matching cost $\hat{\mathbf{C}}^s$ at the $s$-th stage is defined as 
\begin{equation}
\begin{aligned}
\hat{\mathbf{C}}^s_{ij} =& \lambda_{\mathrm{cls}} 
\left( 
\eta_{cls}^{i, s} \hat{\mathcal{C}}_{\mathrm{cls}}(p_i^{s-1},\hat{y}_j)
+ (1- \eta_{cls}^{i, s}) \mathcal{C}_{\mathrm{cls}}(p_i^{s},\hat{y}_j)
\right) \\
& + \lambda_{\mathrm{box}} 
\left( 
\eta_{loc}^{i, s} \hat{\mathcal{C}}_{\mathrm{box}}(b_i^{s-1},\hat{b}_j)
+ (1- \eta_{loc}^{i, s}) \mathcal{C}_{\mathrm{box}}(b_i^{s},\hat{b}_j)
\right) \\
& + \lambda_{\mathrm{iou}} 
\left( 
\eta_{loc}^{i, s} \hat{\mathcal{C}}_{\mathrm{iou}}(b_i^{s-1},\hat{b}_j)
+ (1- \eta_{loc}^{i, s}) \mathcal{C}_{\mathrm{iou}}(b_i^{s},\hat{b}_j)
\right).
\end{aligned}
\end{equation}
Finally, the obtained cost matrices are fed to the Hungarian algorithm for searching the optimal assignments.

By adopting the historical matching states across decoding stages and the query-specified adaptive smoothing factors, our Momentum-based Bipartite Matching strategy provides an effective solution to mitigate inter-stage matching inconsistency, thus offering smooth and stable optimization trajectories for supervising progressive query refinement in the decoder.

\begin{table*}[t]
\centering
\caption{Comparisons with the advanced oriented detectors on DOTA-V1.0 \textsc{Test}. \textcolor{red}{Red} and \textcolor{blue}{blue} indicate top-2 performers. $\dag$ indicates reproduced results using 3x training schedule reported in \cite{arsdetr_cvpr2023}. 
}
\label{tbl:comp_dota_v1_0_single_scale}
\resizebox{1.0\linewidth}{!}{
\setlength{\tabcolsep}{5pt}
\begin{tabular}{l|c|ccccccccccccccc|c} 
    \toprule
    Method & Back. & PL & BD & BR & GTF & SV & LV & SH & TC & BC & ST & SBF & RA & HA & SP & HC & AP$_{50}$  \\ 
    \bottomrule
    \multicolumn{18}{l}{Models with R-50 backbone} \\
    \hline
    
    R-Deform-DETR~\cite{deformable_detr_iclr2020} & R-50 
    & 78.95 & 68.64 & 32.57 & 55.17 & 72.53 & 57.77 & 73.71 & 88.36 & 75.46 & 79.34 & 45.36 & 53.78 & 52.94 & 66.35 & 50.38 & 63.42  \\
    
	Rotated RetinaNet$^\dag$~\cite{lin2017focal} & R-50 
	& 87.33 & 78.91 & 46.45 & 69.81 & 67.72 & 62.34 & 73.59 & 90.85 & 82.79 & 79.37 & 59.62 & 61.89 & 65.01 & 67.76 & 44.95 & 69.23  \\
    AO$^2$-DETR$^\dag$ ~\cite{dai2022ao2detr} & R-50 
	& 87.99 & 79.46 & 45.74 & 66.64 & 78.90 & 73.90 & 73.30 & 90.40 & 80.55 & 85.89 & 55.19 & 63.62 & 51.83 & 70.15 & 60.04 & 70.91  \\

    EMO2-DETR~\cite{hu2023emo2detr}	&    R-50	&	88.08	&	77.91	&	43.17	&	62.91	&	74.01	&	75.09	&	79.21	&	90.88	&	81.50	&	84.04	&	51.92	&	59.44	&	64.74	&	71.81	&	58.96	&	70.91 	\\

	R-Deform-DETR(CSL)$^\dag$~\cite{cls_eccv2020} & R-50 
	& 86.27 & 76.66 & 46.64 & 65.29 & 76.80 & 76.32 & 87.74 & 90.77 & 79.38 & 82.36 & 54.00 & 61.47 & 66.05 & 70.46 & 61.97 & 72.15 \\
    Rotated FCOS$^\dag$~\cite{fcos_iccv2019} & R-50 
	& 88.52 & 77.54 & 47.06 & 63.78 & 80.42 & 80.50 & 87.34 & 90.39 & 77.83 & 84.13 & 55.45 & 65.84 & 66.15 & 72.77 & 49.17 & 72.45  \\
    SASM$^\dag$~\cite{sasm_aaai2022} & R-50 
	& 87.51 & 80.15 &  51.07 & 70.35 &74.95 & 75.80 & 84.23 & 90.90 & 80.87 & 84.93 & 58.51& 65.59& 69.74& 70.18& 42.31 & 72.47 \\
	
    GWD$^\dag$ ~\cite{yang2021rethinkingrotatedgwd} & R-50 
	& 89.34 & 80.07 & 41.94 & 72.58 & 79.76 & 69.93 & 85.38 & 90.76 & 83.16 & 82.38 & 62.75 & 66.06 & 60.82 & 69.39 & 57.84 & 72.81 \\

	PSC$^\dag$~\cite{psc_cvpr2023} & R-50 & 89.65 & 83.80 & 43.64 & 70.98 & 79.00 & 71.35 & 85.08 & 90.90 & 84.28 & 82.51 & 60.64 & 65.06 & 62.52 & 69.61 & 54.00 & 72.87 \\
    Gliding Vertex$^\dag$~\cite{xu2020glidingvertex} & R-50 
	& 88.71 & 77.22 & 52.00 & 70.85 & 73.75 & 74.81 & 86.55 & 90.89 & 80.41 & 84.63 & 57.66 & 62.88 & 68.49 & 71.86 & 58.17 & 73.26  \\
    
	KFIoU$^\dag$ ~\cite{kfiou_iclr2022} & R-50 
	& 89.20 & 76.40 & 51.64 & 70.15 & 78.31 & 76.43 & 87.10 & 90.88 & 81.68 & 82.22 & 64.65 & 64.84 & 66.77 & 70.68 & 49.52 & 73.37 \\
    Rotated ATSS$^\dag$~\cite{atss_cvpr2020} & R-50 
	& 88.94 & 79.89 & 48.71 & 70.74 & 75.80 & 74.02 & 84.14 & 90.89 & 83.19 & 84.05 & 60.48 & 65.06 & 66.74 & 70.14 & 57.78 & 73.37 \\
	
	R$^3$Det$^\dag$~\cite{yang2021r3det} & R-50 
	& 89.24 & 83.32 & 48.03 & 72.52 & 77.52 & 76.72 & 86.48 & 90.89 & 82.33 & 83.51 & 60.96 & 63.09 & 67.58 & 69.27 & 49.50 & 73.40 \\

    KLD$^\dag$~\cite{KLD} 
	& R-50 & 89.08 & 84.18 & 43.77 & 72.33 & 79.85 & 73.58 & 85.69 & 90.88 & 85.14 & 81.96 & 65.86 & 64.60 & 63.60 & 68.26 & 53.19 & 73.46  \\
	
	ARS-DETR$^\dag$ ~\cite{arsdetr_cvpr2023} & R-50 
	& 86.61 & 77.26 & 48.84 & 66.76 & 78.38 & 78.96 & 87.40 & 90.61 & 82.76 & 82.19 & 54.02 & 62.61 & 72.64 & 72.80 & 64.96 & 73.79  \\

	Rotated Faster RCNN$^\dag$~\cite{ren2015fasterrcnn} &	R-50 
	& 89.09 & 78.28 & 48.93 & 71.54 & 74.01 & 74.99 & 85.90 & 90.84 & 86.87 & 85.03 & 57.97 & 69.74 & 68.10 & 71.28 & 56.88 & 73.96  \\
	
	ReDet$^\dag$~\cite{han2021redet} & R-50 
	& 88.94 & 78.07 & 51.19 & 72.76 & 74.26 & 78.08 & 87.44 & 90.84 & 80.79 & 78.59 & 60.85 & 64.22 & 76.84 & 72.79 & 54.85 & 74.03  \\
	
	RoI-Trans$^\dag$~\cite{roi_trans_cvpr2019} & R-50 
	& 89.01 & 77.48 & 51.64 & 72.07 & 74.43 & 77.55 & 87.76 & 90.81 & 79.71 & 85.27 & 58.36 & 64.11 & 76.50 & 71.99 & 54.06 & 74.05  \\

    Oriented Reppoints$^\dag$~\cite{li2022orientedreppoints} & R-50 & 88.52 & 80.62 & 52.68 & 73.04 & 79.61 & 80.73 & 87.76 & 90.89 & 81.82 & 85.33 & 59.95 & 64.88 & 73.81 & 69.84 & 46.18 & 74.38\\
	
	CFA$^\dag$~\cite{cfa_cvpr2021} & R-50 
	& 88.34 & 83.09 & 51.92 & 72.23 & 79.95 & 78.68 & 87.25 & 90.90 & 85.38 & 85.71 & 59.63 & 63.05 & 73.33 & 70.36 & 47.86 & 74.51  \\

    S$^2$A-Net$^\dag$~\cite{s2anet_tgrs2021} & R-50 
    & 89.26 & 84.11 & 51.97 & 72.78 & 78.23 & 79.41 & 87.46 & 90.85 & 85.62 & 84.09 & 60.18 & 65.90 & 72.54 & 71.59 & 55.31 & 75.29  \\
    
    Oriented Former~\cite{zhao2024orientedformer} & R-50 
	& 88.14 & 79.13 & 51.96 & 67.34 & 81.02 & 83.26 & 88.29 & 90.90 & 85.57 & 86.25 & 60.84 & 66.36 & 73.81 & 71.23 & 56.49 & 75.37  \\
    
    Oriented RCNN~\cite{xie2021orientedrcnn} & R-50 
	& 89.35 & 81.41 & 52.71 & 75.03 & 79.04 & 82.41 & 87.82 & 90.90 & 86.40 & 85.30 & 63.36 & 65.70 & 68.28 & 70.48 & 57.23 & 75.69  \\

    ReDiffDet~\cite{zhao2025rediffdet} & R-50 & 85.45 & 77.84 & 50.39 & 71.63 & 80.90 & 84.85 & 88.77 & 90.88 & 87.00 & 86.08 & 63.30 & 63.43 & 76.01 & 76.03 & 60.16 & \textcolor{blue}{76.18} \\
    \hline 

    \rowcolor{gray!30}
    \textbf{Ours} & R-50 &
    86.85 & 82.66 & 54.14 & 75.37 &
    76.69 & 81.91 & 88.83 & 90.87 & 83.12 & 85.50 &
    61.46 & 62.56 & 76.63 & 72.02 & 70.89 &
    \textcolor{red}{76.63} \\  

    \hline

    \multicolumn{18}{l}{Models with stronger backbones} \\
    \hline
    
    EMO2-DETR~\cite{hu2023emo2detr} & Swin-T 
    & 89.03 & 79.59 & 48.71 & 60.23 & 77.34 & 76.42 & 84.53 & 90.77 & 84.80 & 85.68 & 48.86 & 67.55 & 66.32 & 71.54 & 53.49 & 72.32 	\\
    
    ARS-DETR$^\dag$ ~\cite{arsdetr_cvpr2023} & Swin-T 
    & 87.78 & 78.58 & 52.58 & 67.69 & 80.19 & 84.32 & 88.19 & 90.68 & 85.92 & 84.76 & 55.18 & 66.89 & 74.57 & 79.09 & 60.35 & 75.79 	\\
    
    Swin-FEE~\cite{11195847} & Swin-T & 89.2 & 82.3 & 53.6 & 74.6 & 79.0 & 77.8 & 87.7 & 90.9 & 86.5 & 86.1 & 60.4 & 65.7 & 75.3 & 70.1 & 57.4 & 75.80\\

    OrientedFormer~\cite{zhao2024orientedformer} & Swin-T 
    & 88.74 & 78.94 & 53.43 & 72.05 & 81.08 & 84.22 & 88.40 & 90.90 & 86.23 & 86.65 & 61.05 & 63.11 & 75.78 & 73.02 & 54.62 & 75.88 	\\
    
    ReDet~\cite{han2021redet}	&	ReR-50	&	88.79 &	82.64 &	53.97 &	74.00 &	78.13 &	84.06 &	88.04 &	90.89 &	87.78 &	85.75 &	61.76 &	60.39 &	75.96 &	68.07 &	63.59	&	76.25  \\

    RoI Trans$^\dag$ ~\cite{roi_trans_cvpr2019} & Swin-T 
    & 88.44 & 85.53 & 54.56 & 74.55 & 73.43 & 78.39 & 87.64 & 90.88 & 87.23 & 87.11 & 64.25 & 63.27 & 77.93 & 74.10 & 60.03 & 76.49  \\

    Oriented RCNN~\cite{arc_cvpr2023} &	ARC-50	&	89.40 & 82.48 & 55.33 & 73.88 & 79.37 & 84.05 & 88.06 & 90.90 & 86.44 & 84.83 & 63.63 & 70.32 & 74.29 & 71.91 & 65.43	&	77.35  \\

    Oriented Reppoints~\cite{li2022orientedreppoints}	&    Swin-T	&	89.11	&	82.32	&	56.71	&	74.95	&	80.70	&	83.70	&	87.67	&	90.81	&	87.11	&	85.85	&	63.60	&	68.60	&	75.95	&	73.54	&	63.76	&	77.63 	\\
    
    Oriented RCNN~\cite{arc_cvpr2023} &	ARC-101	&
    89.39 &	83.58 &	57.51 &	75.94 &	78.75 &	83.58 &	88.08 &	90.90 &	85.93 &	85.38 &	64.03 &	68.65 &	75.59 &	72.03 &	65.68 &	\textcolor{blue}{77.70} \\
    \hline

    \rowcolor{gray!30}
    \textbf{Ours}  & Swin-T &		
    87.70   &   82.11   &   54.88   &   76.51   &
    78.79   &   84.92   &   88.77   &   90.82   &	
    87.71   &   86.51   &   67.41   &   65.71   &   
    77.48   &   72.67   &   66.52   &	
    \textcolor{red}{78.00} \\
    \bottomrule  
\end{tabular}}
\end{table*}

\subsection{Training IGOFormer}

Our IGOFormer consists of a lightweight proposal network and an Intrinsic Geometry-aware Decoder with $S$ decoding stages, and these network components are jointly end-to-end optimized with the assistance of our Momentum-based Bipartite Matching strategy.
To achieve this goal, we adopt the vanilla bipartite matching for conducting label assignments on the predictions from our proposal network.
In the decoder, each decoding stage produces a fixed-length set of predictions, and, instead of conducting label assignment in a stage-wise manner, we feed these predictions to our Momentum-based Bipartite Matching scheme for providing smooth and stable matching trajectories.

Then, the overall loss is formulated to summarize the classification error $\mathcal{L}_{\mathrm{cls}}$, box difference $\mathcal{L}_{\mathrm{box}}$ and box spatial misalignment $\mathcal{L}_{\mathrm{iou}}$,
\begin{equation}
\label{eq:iorformer_loss}
\mathcal{L} = 
\lambda_{\mathrm{cls}}
\mathcal{L}_{\mathrm{cls}}
+
\lambda_{\mathrm{box}}
\mathcal{L}_{\mathrm{box}}
+
\lambda_{\mathrm{iou}}
\mathcal{L}_{\mathrm{iou}},
\end{equation}
where Smooth-L1 loss and Rotated IoU loss are adopted for $\mathcal{L}_{\mathrm{box}}$ and $\mathcal{L}_{\mathrm{iou}}$, respectively.
The coefficients $\lambda_{\mathrm{cls}}$, $\lambda_{\mathrm{box}}$ and $\lambda_{\mathrm{iou}}$ are introduced to control the contribution of each loss component.

\section{Experiments}

\subsection{Datasets}

Our experimental comparisons are conducted on three challenging datasets, including DOTA-V1.0, DOTA-V1.5 and DIOR-R.

\textbf{DOTA-V1.0 and DOTA-V1.5}~\cite{xia2018dota} are two widely adopted large-scale oriented object detection benchmarks, where objects of interest on each image are featured by their arbitrary orientations and tiny object scales.
DOTA-V1.0 contains 2,806 large aerial images, and the image size ranges from around $800 \times 800$ to $4000 \times 4000$. 
There are 188,282 annotated instances among 15 common categories including Plane (PL), Baseball diamond (BD), Bridge (BR), Ground track field (GTF), Small vehicle (SV), Large vehicle (LV), Ship (SH), Tennis court (TC), Basketball court (BC), Storage tank (ST), Soccer-ball field (SBF), Roundabout (RA), Harbor (HA), Swimming pool (SP), Helicopter (HC). 
DOTA-V1.5 is released with a new category named Container crane (CC), which contains 403,318 annotated object instances in total. 
For experiments on DOTA dataset,  we employ the combined training and validation subsets for training, while reserving the test partition exclusively for performance assessment. We follow the same image patch cropping and augmentation pipeline as \cite{arsdetr_cvpr2023}. The input images are partitioned into $1024\times1024$ pixel patches with a 200-pixel overlap to ensure comprehensive spatial coverage. 

\textbf{DIOR-R} \cite{aopg_tgrs2022} is another comprehensive aerial imagery collection annotated with oriented bounding boxes. 
Comprising 23,463 images with 192,518 annotated instances, this dataset encompasses 20 object categories: Airplane (APL), Airport (APO), Baseball Field (BF), Basketball Court (BC), Bridge (BR), Chimney (CH), Expressway Service Area (ESA), Expressway Toll Station (ETS), Dam (DAM), Golf Field (GF), Ground Track Field (GTF), Harbor (HA), Overpass (OP), Ship (SH), Stadium (STA), Storage Tank (STO), Tennis Court (TC), Train Station (TS), Vehicle (VE), and Windmill (WM). For experimental purposes, we employ the combined training and validation sets for model development, while reserving the test set exclusively for performance evaluation.
For experiments on DIOR-R dataset, we maintain the original 800×800 pixel resolution as the network input. 
To mitigate overfitting risks during the training phase, we implement a minimal data augmentation strategy limited to randomized horizontal, vertical, and diagonal flips, deliberately excluding more complex augmentation techniques.

\subsection{Implementation Details}

Our detector contains a stack of 6 decoding stages, and 500 object queries are selected for each image.
The number of groups $M$ is set to 8. 
During training, the backbone is initialized with pre-trained weights. 
In our Momentum-based Bipartite Matching, the scaling coefficient is set to 0.8, and the cost weights $\lambda_{cls}$, $\lambda_{box}$, and $\lambda_{iou}$ are set to 2.0, 5.0, and 5.0, respectively.
We adopt the AdamW optimizer for training with the ‘$3 \times$’ (36 epochs) schedule on all datasets. 
The initial learning rate is set to $1 \times 10^{-4}$, and the learning rate is further decreased by a factor of 10 at the 24th and 33rd epochs.
All models in this paper are trained on 2 NVIDIA GeForce RTX 3090 GPUs with a total batch size of 8, implemented in the PyTorch-based MMRotate framework~\cite{zhou2022mmrotate}.

\begin{figure*}[t]
\centering
\footnotesize
\begin{tabular}{cccccc}
\includegraphics[width=0.14\linewidth]{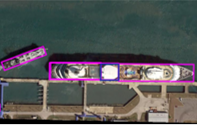}  &
\includegraphics[width=0.14\linewidth]{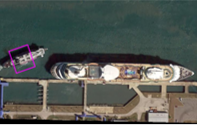}  &
\includegraphics[width=0.14\linewidth]{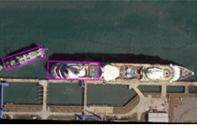}  &
\includegraphics[width=0.14\linewidth]{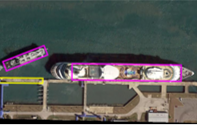}  &
\includegraphics[width=0.14\linewidth]{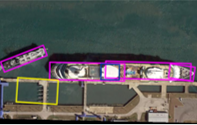}  &
\includegraphics[width=0.14\linewidth]{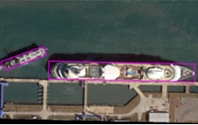}  \\

\includegraphics[width=0.14\linewidth]{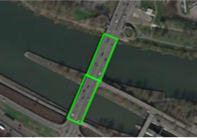}  &
\includegraphics[width=0.14\linewidth]{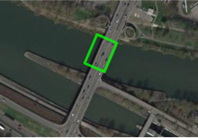}  &
\includegraphics[width=0.14\linewidth]{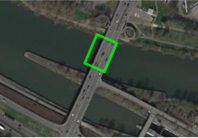}  &
\includegraphics[width=0.14\linewidth]{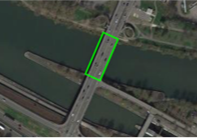}  &
\includegraphics[width=0.14\linewidth]{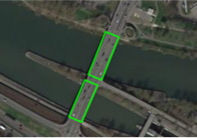}  &
\includegraphics[width=0.14\linewidth]{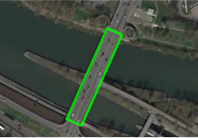}  \\

\includegraphics[width=0.14\linewidth]{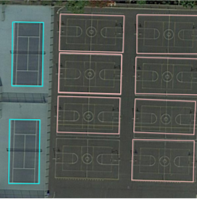}  &
\includegraphics[width=0.14\linewidth]{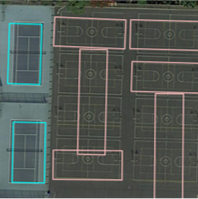}  &
\includegraphics[width=0.14\linewidth]{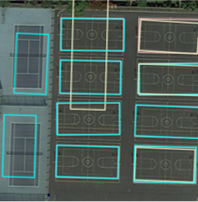}  &
\includegraphics[width=0.14\linewidth]{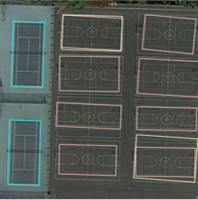}  &
\includegraphics[width=0.14\linewidth]{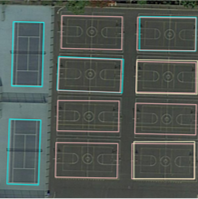}  &
\includegraphics[width=0.14\linewidth]{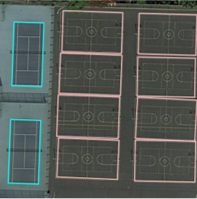}  \\

\includegraphics[width=0.14\linewidth]{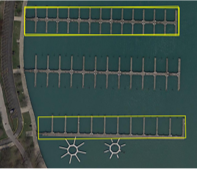}  &
\includegraphics[width=0.14\linewidth]{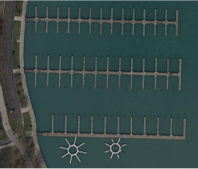}  &
\includegraphics[width=0.14\linewidth]{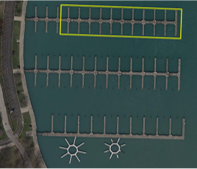}  &
\includegraphics[width=0.14\linewidth]{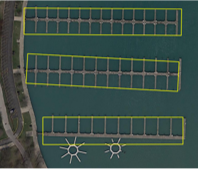}  &
\includegraphics[width=0.14\linewidth]{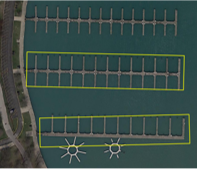}  &
\includegraphics[width=0.14\linewidth]{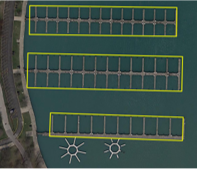}  \\

ARS-DETR & GWD & KLD & Oriented R-CNN & Oriented Reppoints & Ours \\
\end{tabular}
\caption{Qualitative comparison against top-ranking oriented object detectors on DOTA-V1.0 \textsc{TEST} set.}
\label{fig:results-visualize}
\end{figure*}

\begin{table*}[!t]
\centering
\caption{Comparisons with the advanced oriented detectors on DOTA-V1.5 \textsc{Test}. \textcolor{red}{Red} and \textcolor{blue}{blue} indicate top-2 performers.}
\label{tbl:comparison_dotav1_5}
\resizebox{1.0\linewidth}{!}{
\setlength{\tabcolsep}{5pt}
\begin{tabular}{l|c|cccccccccccccccc|c}
\toprule
Method  &   Backbone    &	
PL  &   RD  &   BR  &   GTF &
SV  &   LV  &   SH  &   TC  &
BC  &   ST  &   SBF &   RA  &
HA  &   SP  &   HC  &   CC	&
AP$_{50}$\\
\hline


EMO2-DETR~\cite{hu2023emo2detr}	&	R-50	&	
71.81	&	75.36	&	45.09	&	58.70	&
48.19	&	73.26	&	80.28	&	90.70	&
73.05	&	76.53	&	39.36	&	65.31	&
56.96	&	69.29	&	47.11	&	15.64		&
61.67 	\\

DCL~\cite{dcl_cvpr2021}	&	R-50		&	
80.34	&	74.39	&	44.21	&	62.01	&
50.24	&	72.15	&	78.86	&	89.28	&
74.04	&	67.04	&	45.46	&	69.06	&	
56.17	&	64.06	&	55.14	&	9.27		&
61.98\\


Rotated Faster RCNN~\cite{ren2015fasterrcnn} &	R-50 & 71.89 & 77.64 & 44.45 & 59.87 & 51.28 & 68.98 & 79.37 & 90.78 & 77.38 & 67.50 & 47.75 & 69.72 & 61.22 & 65.28 & 60.47 & 1.54 & 62.00 \\

KLD~\cite{KLD}	&	R-50		&	
79.46	&	76.49	&	41.98	&	65.28	&
50.43	&	69.91	&	82.09	&	90.35	&	
74.68	&	58.89	&	51.57	&	66.11	&	
64.07	&	64.04	&	53.33	&	11.24	&
62.50	\\

Mask RCNN~\cite{he2017maskrcnn} &	R-50 & 76.84 & 73.51 & 49.90 & 57.80 & 51.31 & 71.34 & 79.75 & 90.46 & 74.21 & 66.07 & 46.21 & 70.61 & 63.07 & 64.46 & 57.81 & 9.42 & 62.67 \\


R$^3$Det~\cite{yang2021r3det}	&	R-50		&	
80.22	&	75.55	&	42.36	&	65.21	&
50.22	&	70.48	&	79.58	&	89.51	&	
75.08	&	66.27	&	54.11	&	69.28	&
55.37	&	66.13	&	57.31	&	9.94		&
62.91	\\

HTC~\cite{chen2019hybrid} &	R-50 & 77.80 & 73.67 &51.40 & 63.99 & 51.54 & 73.31 & 80.31 & 90.48 & 75.12 & 67.34 & 48.51 & 70.63 & 64.84 & 64.48 & 55.87 & 5.15 & 63.40 \\

AO$^2$-DETR~\cite{dai2022ao2detr}	&	R-50		&	
79.55	&	78.14	&	42.41	&	61.23	&
55.34	&	74.50	&	79.57	&	90.64	&	
74.76	&	77.58	&	53.56	&	66.91	&
58.56	&	73.11	&	69.64	&	24.71	&
66.26	\\

ReDet~\cite{han2021redet}	&	ReR-50	&	79.20 &	82.81 &	51.92 &	71.41 &	52.38 &	75.73 &	80.92 &	90.83 &	75.81 &	68.64 &	49.29 &	72.03 &	73.36 &	70.55 &	63.33 &	11.53	&	\textcolor{blue}{66.86}	\\
\hline

\rowcolor{gray!30}
\textbf{Ours} & R-50 & 71.99 & 78.11 & 51.74 & 70.03 & 58.46 & 77.38 & 88.61 & 90.84 & 79.20 & 78.66 & 58.82 & 67.74 & 73.70 & 72.25 & 61.28 & 23.24 & \textcolor{red}{68.88} \\

\bottomrule
\end{tabular}}
\end{table*}

\begin{table*}[t]
\centering
\caption{Comparisons with the advanced oriented detectors on DIOR-R. \textcolor{red}{Red} and \textcolor{blue}{blue} indicate top-2 performers.}
\label{tbl:comparison_dior}
\resizebox{1.0\linewidth}{!}{
\setlength{\tabcolsep}{5pt}
\begin{tabular}{l|c|c|c|c|c|c|c|c|c|c|c|c|c|c|c|c|c|c|c|c|c} 
\toprule
Method & APL & APO & BF & BC & BR & CH & ESA & ETS & DAM & GF & GTF & HA & OP & SH & STA & STO & TC & TS & VE & WM & AP$_{50}$ \\ 
\hline
SASM~\cite{sasm_aaai2022} & 61.41 & 46.03 & 73.22 & 82.04 & 29.41 & 71.03 & 69.22 & 53.91 & 30.63 & 70.04 & 77.02 & 39.33 & 47.51 & 78.62 & 66.14 & 62.92 & 79.93 & 54.41 & 40.62 & 63.01 & 59.81 \\
GWD~\cite{yang2021rethinkingrotatedgwd} & 69.68 & 28.83 & 74.32 & 81.49 & 29.62 & 72.67 & 76.45 & 63.14 & 27.13 & 77.19 & 78.94 & 39.11 & 42.18 & 79.10 & 70.41 & 58.69 & 81.52 & 47.78 & 44.47 & 62.63 & 60.31 \\
R$^3$Det~\cite{yang2021r3det} & 62.55 & 43.44 & 71.72 & 81.48 & 36.49 & 72.63 & 79.50 & 64.41 & 27.02 & 77.36 & 77.17 & 40.53 & 53.33 & 79.66 & 69.22 & 61.10 & 81.54 & 52.18 & 43.57 & 64.13 & 61.91 \\
Gliding Vertex~\cite{xu2020glidingvertex} & 62.67 & 38.56 & 71.94 & 81.20 & 37.73 & 72.48 & 78.62 & 69.04 & 22.81 & 77.89 & 82.13 & 46.22 & 54.76 & 81.03 & 74.88 & 62.54 & 81.41 & 54.25 & 43.22 & 65.13 & 62.91 \\
Rotated FCOS~\cite{fcos_iccv2019} & 62.31 & 42.18 & 75.34 & 81.32 & 39.26 & 74.89 & 77.42 & 68.67 & 26.00 & 73.94 & 78.73 & 41.28 & 54.19 & 80.61 & 66.92 & 69.17 & 87.20 & 52.31 & 47.08 & 65.21 & 63.21 \\
Rotated Faster RCNN~\cite{ren2015fasterrcnn} & 63.07 & 40.22 & 71.89 & 81.36 & 39.67 & 72.51 & 79.19 & 69.45 & 26.00 & 77.93 & 82.28 & 46.91 & 53.90 & 81.03 & 75.77 & 62.54 & 81.42 & 54.50 & 43.17 & 65.73 & 63.41 \\
Rotated ATSS~\cite{atss_cvpr2020} & 62.19 & 44.63 & 71.55 & 81.42 & 41.08 & 72.37 & 78.54 & 67.50 & 30.56 & 75.69 & 79.11 & 42.77 & 56.31 & 80.92 & 67.78 & 69.24 & 81.62 & 55.45 & 47.79 & 64.10 & 63.52 \\
ReDet~\cite{han2021redet} & 63.22 & 44.18 & 72.11 & 81.26 & 43.83 & 72.72 & 79.10 & 69.78 & 28.45 & 78.69 & 77.18 & 48.24 & 56.81 & 81.17 & 69.17 & 62.73 & 81.42 & 54.90 & 44.04 & 66.37 & 63.81 \\
RoI Trans.~\cite{roi_trans_cvpr2019} & 63.18 & 44.33 & 71.91 & 81.26 & 42.19 & 72.64 & 79.30 & 69.67 & 29.42 & 77.33 & 82.88 & 48.09 & 57.03 & 81.18 & 77.32 & 62.45 & 81.38 & 54.34 & 43.91 & 66.30 & 64.31 \\
AOPG~\cite{aopg_tgrs2022}	&	62.39	&	37.79	&	71.69	&	87.63	&	40.90	&	72.47	&	31.08	&	65.42	&	77.99	&	73.20	&	81.94	&	42.32	&	54.45	&	81.17	&	72.69	&	71.31	&	81.49	&	60.04	&	52.38	&	69.99	&	64.41	\\
S$^2$A-Net~\cite{s2anet_tgrs2021} & 67.98 & 44.44 & 71.63 & 81.39 & 42.66 & 72.72 & 79.03 & 70.40 & 27.08 & 75.56 & 81.02 & 43.41 & 56.45 & 81.12 & 68.00 & 70.03 & 87.07 & 53.88 & 51.12 & 65.31 & 64.50\\
Oriented RCNN~\cite{xie2021orientedrcnn} & 63.31 & 43.10 & 71.89 & 81.17 & 44.78 & 72.64 & 80.12 & 69.67 & 33.78 & 77.92 & 83.11 & 46.29 & 58.31 & 81.17 & 74.54 & 62.32 & 81.29 & 56.30 & 43.78 & 65.26 & 64.53  \\ 
KLD~\cite{KLD} & 66.52 & 46.80 & 71.76 & 81.43 & 40.81 & 78.25 & 79.23 & 66.63 & 29.01 & 78.68 & 80.19 & 44.88 & 57.23 & 80.91 & 74.17 & 68.02 & 81.48 & 54.63 & 47.80 & 64.41 & 64.63 \\
CFA~\cite{cfa_cvpr2021} & 61.10 & 44.93 & 77.62 & 84.67 & 37.69 & 75.71 & 82.68 & 72.03 & 33.41 & 77.25 & 79.94 & 46.20 & 54.27 & 87.01 & 70.43 & 69.58 & 81.55 & 55.51 & 49.53 & 64.92 & 65.25 \\
ARS-DETR~\cite{arsdetr_cvpr2023}	& 65.82 & 53.40 & 74.22 & 81.11 & 42.13 & 76.23 & 82.24 & 71.52 & 38.90 & 75.91 & 77.91 & 33.03 & 57.02 & 84.82 & 69.71 & 72.20 & 80.33 & 58.91 & 51.52 & 70.73 & 65.90  \\
OrientedFormer~\cite{zhao2024orientedformer}	& 65.65 & 48.69 & 78.79 & 87.17 & 41.90 & 76.34 & 34.37 & 72.14 & 81.40 & 75.34 & 79.83 & 45.15 & 56.12 & 88.66 & 67.59 & 72.68 & 87.32 & 60.31 & 56.54 & 69.56 & \textcolor{blue}{67.28}  \\
\hline

\rowcolor{gray!30}
Ours
&	73.98   &	49.79	&	75.85	&	87.07	&   49.51
&	78.65	&	84.51	&	72.87	&	36.22	&	73.55	
&	82.66	&	47.48	&	60.90	&	89.61	&	76.76
&	76.26	&	87.38	&	52.71	&	58.33	&	71.25
&	\textcolor{red}{69.27}	\\

\bottomrule
\end{tabular}}
\end{table*}

\begin{table*}[!t]
  \centering
  \footnotesize
  \caption{Effect of each module. IGA: Intrinsic Geometry Augmentation. MG: Multi-group Scheme. MBM: Momentum-based Bipartite Matching.}
  \label{tab:ablation_comp}
  \begin{tabular}{ccc|l}
    \toprule
    IGA & MG & MBM & AP\textsubscript{50} (\%)\\
    \midrule
    \XSolidBrush & \XSolidBrush & \XSolidBrush & 62.51 \\
    \Checkmark & \XSolidBrush & \XSolidBrush & 69.73 (+7.22)\\
    \XSolidBrush & \Checkmark & \XSolidBrush & 65.28 (+2.77)\\
    \XSolidBrush & \XSolidBrush & \Checkmark & 66.70 (+4.19)\\
    \bottomrule
  \end{tabular}
  \hspace{0.5cm}
  \begin{tabular}{ccc|l}
    \toprule
    IGA & MG & MBM & AP\textsubscript{50} (\%) \\
    \midrule
    \XSolidBrush & \XSolidBrush & \XSolidBrush & 62.51 \\
    \Checkmark & \XSolidBrush & \XSolidBrush & 69.73 (+7.22) \\
    \Checkmark & \Checkmark & \XSolidBrush & 69.90 (+7.39)\\
    \Checkmark & \Checkmark & \Checkmark & 71.44 (+8.93)\\
    \bottomrule
  \end{tabular}
  \hspace{0.5cm}
  \begin{tabular}{ccc|l}
    \toprule
    IGA & MG & MBM & AP\textsubscript{50} (\%) \\
    \midrule
    \XSolidBrush & \Checkmark & \Checkmark & 69.39 (-2.05) \\
    \Checkmark & \XSolidBrush & \Checkmark & 70.52 (-0.92)\\
    \Checkmark & \Checkmark & \XSolidBrush & 69.90 (-1.54)\\
    \Checkmark & \Checkmark & \Checkmark & 71.44 \\
    \bottomrule
  \end{tabular}
\end{table*}

\subsection{Main Results}


\textbf{Results on DOTA-V1.0.} 
We first compare the results of our IGOFormer with R-50 backbone against a list of top performers under single-scale training and inference.
As presented in \cref{tbl:comp_dota_v1_0_single_scale}, our IGOFormer achieves an AP$_{50}$ of 76.63\%, which outperforms most known oriented object detectors with the same backbone. 
Compared with known oriented detection transformers, our IGOFormer outperforms OrientedFormer and ARS-DETR by 1.26\% and 2.84\% with the same R-50 backbone, respectively.
With a stronger Swin-T backbone\cite{liu2021swintransformer}, we achieve an AP$_{50}$ of 78.00\%. 
Compared to the best performing method using the ARC-101 backbone, our approach demonstrate superior performance. 
This implies that by the Intrinsic Geometry-aware decoding module, and Momentum-based Bipartite Matching, our IGOFormer benefits from enhanced geometric information extraction capabilities and demonstrates improved inter-stage matching stability.

\cref{fig:results-visualize} presents a qualitative comparison between IGOFormer and state-of-the-art oriented object detection methods. 
The results demonstrate that while existing approaches frequently exhibit critical limitations such as missed detections, false positives, and localization inaccuracies, the proposed IGOFormer overcomes these challenges through its Intrinsic Geometry-aware decoding module. 
This innovative component enables precise extraction of objects' geometric structural features, leading to significantly improved detection performance. 
The visual comparison highlights IGOFormer's superior capability in maintaining structural integrity and positional accuracy, particularly for complex oriented object arrangements.

\textbf{Results on DOTA-V1.5.} 
DOTA-V1.5 includes a greater number of extremely small objects compared to DOTA-V1.0. \cref{tbl:comparison_dotav1_5} shows the comparison results between IGOFormer and the recent state-of-the-art methods in DOTA-V1.5 dataset, our IGOFormer achieves the highest AP$_{50}$ of 68.88\% among the detectors with R-50 backbone in single-scale training and inference. 
Notably, IGOFormer outperforms existing oriented detection transformer methods, achieving performance improvements of 2.62\% and 7.11\% over AO$^2$-DETR and EMO2-DETR, respectively, with the same R-50 backbone. Additionally, when compared to the state-of-the-art method ReDet, which uses the ReR-50 backbone, our approach demonstrates a performance gain of 2.02\%.

\textbf{Results on DIOR-R.}
Results on DIOR-R dataset are shown in \cref{tbl:comparison_dior}, our IGOFormer with R-50 backbone achieves an AP$_{50}$ of 69.27\%, and surpasses the known top performer OrientedFormer by 1.99\%.
In particular, our method achieves superior performance in categories such as airplane, bridge, and vehicle, benefiting from the proposed components in effectively capturing the geometric features of the objects while significantly improving the stability of inter-stage matching.

\subsection{Ablation and Analysis}

To validate the effectiveness of the proposed method, we conducted ablation studies on the DOTA-V1.0 benchmark dataset. 
The ablation experiments employed the \textsc{Train} set for model training and utilized the \textsc{Val} set for performance evaluation. All ablation experiments were conducted with models using R-50 backbone.

\textbf{Ablation on Each Component.} 
We build a baseline model based on Sparse RCNN using R-50 backbone, by introducing a proposal network and additional angle prediction head, which achieves an AP$_{50}$ of 62.51\% on DOTA-V1.0 \textsc{Val} set.
As summarized in \cref{tab:ablation_comp}, by integrating our Intrinsic Geometry Augmentation scheme, the AP$_{50}$ score is boosted by 7.22\%, implying the importance of geometric information for tiny and oriented object detection. 
When our Momentum-based Bipartite Matching is applied to substitute the stage-wise bipartite matching in the baseline model, it achieves a performance gain of 4.19\% from 62.51\% in AP$_{50}$, indicating that maintaining the inter-stage matching consistency plays an important role in high-precision tiny and oriented object detection.

\textbf{Ablation on Combining Components.} 
As summarized in \cref{tab:ablation_comp}, by progressively enabling our model components, the detection accuracy is improved consistently.
Enabling our Intrinsic Geometry Augmentation and Multi-group Scheme, the AP$_{50}$ is improved by 7.39\%.
Another gain of 1.54\% is achieved by further activating our Momentum-based Bipartite Matching scheme to approach an AP$_{50}$ of 71.44\%.
Then we validate the contribution of each component by removing it individually.
Notably, the absence of Intrinsic Geometry Augmentation (IGA) results in a 2.05\% performance decrease (69.39\% vs 71.44\%), underscoring the value of geometry-aware representations. 
When disabling the Momentum-based Bipartite Matching (MBM) while retaining other components, the performance drops from 71.44\% to 69.90\%. 
These consistent performance gaps validate the importance of the proposed Intrinsic Geometry-aware decoding module and Momentum-based Bipartite Matching scheme by obtaining Intrinsic Geometry-aware representation and enhancing the inter-stage matching stability.

\textbf{Design of Intrinsic Geometry-aware Decoding Module.} 
We compare different designs of the Intrinsic Geometry-aware decoding module on the DOTA-V1.0 \textsc{Val} set, as shown in \cref{decoding-design}.
Our default design is to first apply the Multi-group dynamic interactive layer followed by our Intrinsic Geometry Augmentation layer, which achieves the highest AP$_{50}$ of 71.44\%.
To validate this design, we conduct four extra ablation experiments. 
The performance significantly decreases when using a reversed structure or double Multi-group dynamic interactive layers. Reversed structure drops performance to 70.92\% (-0.52\%), while double MDI layers further degrade results to 70.08\% (-1.36\%).

\begin{table}[t]
  \centering
  \caption{Design of our decoding module.}
  \begin{tabular}{cc|c}
    \toprule
    Decoding Method 1 & Decoding Method 2 & AP\textsubscript{50} (\%) \\
    \midrule
    MDI & MDI & 70.08\\
    MDI & IGA & 71.44 \\
    IGA & MDI & 70.92 \\
    \bottomrule
  \end{tabular}%
  \label{decoding-design}%
\end{table}%

\textbf{Analysis on Number of Groups $M$.}  
As demonstrated in \cref{m-values-impact}, we systematically investigated the influence of varying group numbers ($M$) on detection performance, evaluated on the DOTA-V1.0 validation set, while simultaneously quantifying the computational complexity in terms of both parameter count and floating-point operations (FLOPs).
The results indicate that the choice of group numbers affects performance by the diversity of feature channel representations and computational complexity. 
We choose M from $\{1,\ 2,\ 4,\ 8,\ 16,\ 32\}$ to systematically analyze this impact. Initially, as the number of groups $M$ increases from 1, AP$_{50}$ improves and approaches the highest AP$_{50}$ result of 71.44\% when $M=8$. 
When the number of groups continues to increase, AP$_{50}$ starts to decrease. 
At the same time, the computational cost decreases as the number of groups increases.
In this work, we set $M=8$ as an optimal balance for maintaining acceptable computational cost while improving detection accuracy.

\begin{table}[t]
  \centering
  \caption{Effect of the number of groups $M$ on detection performance and computational cost.}
  \begin{tabular}{c|cccccc}
    \toprule
    $M$ & 1 & 2 & 4 & 8 & 16 & 32 \\
    \midrule
    AP\textsubscript{50} (\%) & 70.52 & 70.00 & 70.84 & 71.44 & 69.94 & 70.06 \\
    FLOPs(G) & 314 & 245 & 211 & 193 & 184 & 72\\
    Param(M) & 259 & 157 & 106 & 81.06 & 68.41 & 62.09 \\
    \bottomrule
  \end{tabular}%
  \label{m-values-impact}%
\end{table}%

\begin{table}[t]
  \centering
  \caption{Performance Comparison of Different Smoothing Strategies in Momentum-based Bipartite Matching.}
  \begin{tabular}{l|c}
    \toprule
    Smooth Factor & AP\textsubscript{50} (\%) \\
    \midrule
    w.o. smoothing & 69.90 \\
    w. fixed factors & 70.38 \\
    w. adaptive factors & 71.44 \\
    \bottomrule
  \end{tabular}%
  \label{smooth-factor-comparison}%
\end{table}%

\textbf{Design of Momentum-based Bipartite Matching.}
We examine the effect of adaptive smoothing factors in the Momentum-based Bipartite Matching scheme on the DOTA-V1.0 \textsc{Val} set. 
As summarized in \cref{smooth-factor-comparison}, the absence of smoothing factors leads to the worst performance, highlighting instability in the matching process. 
When combining both the adaptive classification and position smoothing factors, the AP$_{50}$ increases to 71.44\%, an improvement of 1.54\% over no smoothing factors and 1.06\% over the fixed smoothing factors. These results clearly demonstrate the significant advantage of adaptive smoothing factors in dynamically adjusting matching costs. In particular, the combination of classification and position smoothing factors enables a more effective global optimization mechanism, substantially enhancing inter-stage matching consistency and significantly improving detection performance.

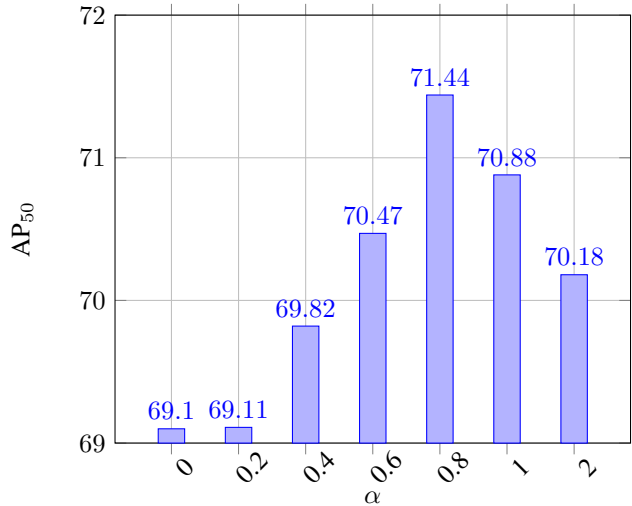
\begin{figure}[t]
    \centering
    \begin{tikzpicture}
        \begin{axis}[
            ybar=2pt, 
            symbolic x coords={0, 0.2, 0.4, 0.6, 0.8, 1, 2}, 
            xtick=data, 
            ylabel={AP$_{50}$}, 
            xlabel={$ \alpha $}, 
            ymin=69, ymax=72, 
            xticklabel style={rotate=45, anchor=north}, 
            grid=major, 
            enlarge x limits={abs=0.75cm}, 
            nodes near coords, 
        ]
        \addplot coordinates {(0, 69.10) (0.2, 69.11) (0.4, 69.82) (0.6, 70.47) (0.8, 71.44) (1, 70.88) (2, 70.18)};
        \end{axis}
    \end{tikzpicture}
    \caption{Study on the impact of $\alpha$ in Momentum-based Bipartite Matching.}
    \label{fig:MBM hyperparameter}
\end{figure}

\textbf{Analysis on Hyperparameter $\alpha$.} 
As shown in \cref{fig:MBM hyperparameter}, we investigated the impact of the hyperparameter $\alpha$ on detection accuracy in Momentum-based Bipartite Matching scheme using the DOTA-V1.0 \textsc{Val} set. To comprehensively evaluate the effectiveness of our method, we conduct experiments across $\alpha$ thresholds $\{0,\, 0.2,\, 0.4,\, 0.6,\, 0.8,\, 1,\, 2\}$. The experimental results indicate that the AP$_{50}$ achieves its best performance when $\alpha$ is set to 0.8, demonstrating that this configuration provides the optimal balance during matching. 
Specifically, when $\alpha$ is set to 0, meaning no smoothing factor is applied, AP$_{50}$ is 69.10\%. Gradually increasing $\alpha$ to 0.8 results in a 2.34\% improvement in AP$_{50}$, demonstrating that a modest increase in $\alpha$ effectively reduces matching instability. However, further increases in $\alpha$ lead to a decline in performance, suggesting that a balanced $\alpha$ value optimizes inter-stage matching consistency.

\subsection{Computational Efficiency}

In our IGOFormer, we introduce Multi-group Scheme for mitigating the computational burden brought by generating query-conditioned filters and pairwise correlation quantification in our decoding module.
As summarized in \cref{tab:comp_efficiency}, compared with query-based detectors Sparse RCNN and ARS-DETR, our model requires slightly increased number of floating operations, yet achieves significantly improved AP$_{50}$.
This demonstrates that our design enhances detection accuracy without imposing an undue computational burden.

\begin{table}[t]
\centering
\caption{Computational efficiency on DOTA-V1.0 \textsc{Test}.}
\begin{tabular}{c|ccccc}
\toprule
Method & AP$_{50}$ & Total FLOPs(G) & Total Param(M)\\ \hline
Sparse R-CNN & 70.96 & 202 & 105.00\\ 
ARS-DETR  & 73.79 & 206 & 41.56\\
Ours & 76.63 & 193 & 81.06\\ \bottomrule
\end{tabular}
\label{tab:comp_efficiency}
\end{table}

\section{Conclusion}
This paper presents IGOFormer, a query-based aerial oriented object detector. 
Our Intrinsic Geometry-aware decoding module explores geometric correlations during feature decoding by encoding object feature relationships into complementary geometric embeddings, which enhances feature representation and improves orientation estimation accuracy. 
Meanwhile, our Momentum-based Bipartite Matching scheme adaptively incorporates matching costs from preceding stages, ensuring matching stability across stages. 
Extensive experiments on three challenging datasets demonstrate the effectiveness of our detector.
We believe this Intrinsic Geometry-aware enhancement, along with the Momentum-based Bipartite Matching scheme, will provide valuable insights for future research in tiny and oriented object detection.

\small
\bibliographystyle{IEEEtran}
\bibliography{references}

\begin{thebibliography}{10}
\providecommand{\url}[1]{#1}
\csname url@samestyle\endcsname
\providecommand{\newblock}{\relax}
\providecommand{\bibinfo}[2]{#2}
\providecommand{\BIBentrySTDinterwordspacing}{\spaceskip=0pt\relax}
\providecommand{\BIBentryALTinterwordstretchfactor}{4}
\providecommand{\BIBentryALTinterwordspacing}{\spaceskip=\fontdimen2\font plus
\BIBentryALTinterwordstretchfactor\fontdimen3\font minus \fontdimen4\font\relax}
\providecommand{\BIBforeignlanguage}[2]{{%
\expandafter\ifx\csname l@#1\endcsname\relax
\typeout{** WARNING: IEEEtran.bst: No hyphenation pattern has been}%
\typeout{** loaded for the language `#1'. Using the pattern for}%
\typeout{** the default language instead.}%
\else
\language=\csname l@#1\endcsname
\fi
#2}}
\providecommand{\BIBdecl}{\relax}
\BIBdecl

\bibitem{shi2018face}
X.~Shi, S.~Shan, M.~Kan, S.~Wu, and X.~Chen, ``Real-time rotation-invariant face detection with progressive calibration networks,'' in \emph{Proceedings of the IEEE Conference on Computer Vision and Pattern Recognition}, 2018, pp. 2295--2303.

\bibitem{kaewkorn2026face}
H.~Kaewkorn, L.~Zhou, W.~Li, and C.~Long, ``Topology-guided semantic face center estimation for rotation-invariant face detection,'' \emph{IEEE Transactions on Image Processing}, 2026.

\bibitem{xia2018dota}
G.-S. Xia, X.~Bai, J.~Ding, Z.~Zhu, S.~Belongie, J.~Luo, M.~Datcu, M.~Pelillo, and L.~Zhang, ``Dota: A large-scale dataset for object detection in aerial images,'' in \emph{Proceedings of the IEEE/CVF Conference on Computer Vision and Pattern Recognition}, 2018, pp. 3974--3983.

\bibitem{fsd_tpami_2025}
J.~Liu, Z.~Lu, Y.~Cen, H.~Hu, Z.~Shao, Y.~Hong, M.~Jiang, and M.~Xu, ``Enhancing object detection with fourier series,'' \emph{IEEE Transactions on Pattern Analysis and Machine Intelligence}, 2025.

\bibitem{ren2015fasterrcnn}
S.~Ren, K.~He, R.~Girshick, and J.~Sun, ``Faster r-cnn: Towards real-time object detection with region proposal networks,'' \emph{Advances in Neural Information Processing Systems}, vol.~28, 2015.

\bibitem{fcos_iccv2019}
Z.~Tian, C.~Shen, H.~Chen, and T.~He, ``Fcos: Fully convolutional one-stage object detection,'' in \emph{Proceedings of the IEEE/CVF International Conference on Computer Vision}, 2019, pp. 9627--9636.

\bibitem{ssd_eccv2016}
W.~Liu, D.~Anguelov, D.~Erhan, C.~Szegedy, S.~Reed, C.-Y. Fu, and A.~C. Berg, ``Ssd: Single shot multibox detector,'' in \emph{Computer Vision--ECCV 2016: 14th European Conference, Amsterdam, The Netherlands, October 11--14, 2016, Proceedings, Part I 14}.\hskip 1em plus 0.5em minus 0.4em\relax Springer, 2016, pp. 21--37.

\bibitem{carion2020endtoenddetr}
N.~Carion, F.~Massa, G.~Synnaeve, N.~Usunier, A.~Kirillov, and S.~Zagoruyko, ``End-to-end object detection with transformers,'' in \emph{European Conference on Computer Vision}.\hskip 1em plus 0.5em minus 0.4em\relax Springer, 2020, pp. 213--229.

\bibitem{deformable_detr_iclr2020}
X.~Zhu, W.~Su, L.~Lu, B.~Li, X.~Wang, and J.~Dai, ``Deformable detr: Deformable transformers for end-to-end object detection,'' in \emph{International Conference on Learning Representations}, 2020.

\bibitem{liu2022dabdetr}
S.~Liu, F.~Li, H.~Zhang, X.~Yang, X.~Qi, H.~Su, J.~Zhu, and L.~Zhang, ``{DAB}-{DETR}: Dynamic anchor boxes are better queries for {DETR},'' 2022.

\bibitem{gao2022adamixer}
Z.~Gao, L.~Wang, B.~Han, and S.~Guo, ``Adamixer: A fast-converging query-based object detector,'' in \emph{Proceedings of the IEEE/CVF Conference on Computer Vision and Pattern Recognition}, 2022, pp. 5364--5373.

\bibitem{arsdetr_cvpr2023}
Y.~Zeng, Y.~Chen, X.~Yang, Q.~Li, and J.~Yan, ``Ars-detr: Aspect ratio-sensitive detection transformer for aerial oriented object detection,'' \emph{IEEE Transactions on Geoscience and Remote Sensing}, vol.~62, pp. 1--15, 2024.

\bibitem{dai2022ao2detr}
L.~Dai, H.~Liu, H.~Tang, Z.~Wu, and P.~Song, ``Ao2-detr: Arbitrary-oriented object detection transformer,'' \emph{IEEE Transactions on Circuits and Systems for Video Technology}, 2022.

\bibitem{zhao2024orientedformer}
J.~Zhao, Z.~Ding, Y.~Zhou, H.~Zhu, W.-L. Du, R.~Yao, and A.~El~Saddik, ``Orientedformer: An end-to-end transformer-based oriented object detector in remote sensing images,'' \emph{IEEE Transactions on Geoscience and Remote Sensing}, 2024.

\bibitem{zhang2022dino}
H.~Zhang, F.~Li, S.~Liu, L.~Zhang, H.~Su, J.~Zhu, L.~M. Ni, and H.-Y. Shum, ``Dino: Detr with improved denoising anchor boxes for end-to-end object detection,'' in \emph{International Conference on Learning Representations}, 2023.

\bibitem{liu2023stable-dino}
S.~Liu, T.~Ren, J.~Chen, Z.~Zeng, H.~Zhang, F.~Li, H.~Li, J.~Huang, H.~Su, J.~Zhu \emph{et~al.}, ``Detection transformer with stable matching,'' in \emph{Proceedings of the IEEE/CVF International Conference on Computer Vision}, 2023, pp. 6491--6500.

\bibitem{chen2023group}
Q.~Chen, X.~Chen, J.~Wang, S.~Zhang, K.~Yao, H.~Feng, J.~Han, E.~Ding, G.~Zeng, and J.~Wang, ``Group detr: Fast detr training with group-wise one-to-many assignment,'' in \emph{Proceedings of the IEEE/CVF International Conference on Computer Vision}, 2023, pp. 6633--6642.

\bibitem{jia2023detrs}
D.~Jia, Y.~Yuan, H.~He, X.~Wu, H.~Yu, W.~Lin, L.~Sun, C.~Zhang, and H.~Hu, ``Detrs with hybrid matching,'' in \emph{Proceedings of the IEEE/CVF conference on computer vision and pattern recognition}, 2023, pp. 19\,702--19\,712.

\bibitem{aopg_tgrs2022}
G.~Cheng, J.~Wang, K.~Li, X.~Xie, C.~Lang, Y.~Yao, and J.~Han, ``Anchor-free oriented proposal generator for object detection,'' \emph{IEEE Transactions on Geoscience and Remote Sensing}, vol.~60, pp. 1--11, 2022.

\bibitem{wang2022anchordetr}
Y.~Wang, X.~Zhang, T.~Yang, and J.~Sun, ``Anchor detr: Query design for transformer-based detector,'' in \emph{Proceedings of the AAAI Conference on Artificial Intelligence}, vol.~36, no.~3, 2022, pp. 2567--2575.

\bibitem{meng2021conditionaldetr}
D.~Meng, X.~Chen, Z.~Fan, G.~Zeng, H.~Li, Y.~Yuan, L.~Sun, and J.~Wang, ``Conditional detr for fast training convergence,'' in \emph{Proceedings of the IEEE/CVF International Conference on Computer Vision}, 2021, pp. 3651--3660.

\bibitem{sun2021sparsercnn}
P.~Sun, R.~Zhang, Y.~Jiang, T.~Kong, C.~Xu, W.~Zhan, M.~Tomizuka, L.~Li, Z.~Yuan, C.~Wang \emph{et~al.}, ``Sparse r-cnn: End-to-end object detection with learnable proposals,'' in \emph{Proceedings of the IEEE/CVF Conference on Computer Vision and Pattern Recognition}, 2021, pp. 14\,454--14\,463.

\bibitem{fair1m_isprs_2022}
X.~Sun, P.~Wang, Z.~Yan, F.~Xu, R.~Wang, W.~Diao, J.~Chen, J.~Li, Y.~Feng, T.~Xu \emph{et~al.}, ``Fair1m: A benchmark dataset for fine-grained object recognition in high-resolution remote sensing imagery,'' \emph{ISPRS Journal of Photogrammetry and Remote Sensing}, vol. 184, pp. 116--130, 2022.

\bibitem{s2anet_tgrs2021}
J.~Han, J.~Ding, J.~Li, and G.-S. Xia, ``Align deep features for oriented object detection,'' \emph{IEEE Transactions on Geoscience and Remote Sensing}, vol.~60, pp. 1--11, 2021.

\bibitem{bbavector_wacv2021}
J.~Yi, P.~Wu, B.~Liu, Q.~Huang, H.~Qu, and D.~Metaxas, ``Oriented object detection in aerial images with box boundary-aware vectors,'' in \emph{Proceedings of the IEEE/CVF Winter Conference on Applications of Computer Vision}, 2021, pp. 2150--2159.

\bibitem{roi_trans_cvpr2019}
J.~Ding, N.~Xue, Y.~Long, G.-S. Xia, and Q.~Lu, ``Learning roi transformer for oriented object detection in aerial images,'' in \emph{Proceedings of the IEEE/CVF Conference on Computer Vision and Pattern Recognition}, 2019, pp. 2849--2858.

\bibitem{xie2021orientedrcnn}
X.~Xie, G.~Cheng, J.~Wang, X.~Yao, and J.~Han, ``Oriented r-cnn for object detection,'' in \emph{Proceedings of the IEEE/CVF International Conference on Computer Vision}, 2021, pp. 3520--3529.

\bibitem{cls_eccv2020}
X.~Yang and J.~Yan, ``Arbitrary-oriented object detection with circular smooth label,'' in \emph{European Conference on Computer Vision}.\hskip 1em plus 0.5em minus 0.4em\relax Springer, 2020, pp. 677--694.

\bibitem{dcl_cvpr2021}
X.~Yang, L.~Hou, Y.~Zhou, W.~Wang, and J.~Yan, ``Dense label encoding for boundary discontinuity free rotation detection,'' in \emph{Proceedings of the IEEE/CVF Conference on Computer Vision and Pattern Recognition}, 2021, pp. 15\,819--15\,829.

\bibitem{yang2021rethinkingrotatedgwd}
X.~Yang, J.~Yan, Q.~Ming, W.~Wang, X.~Zhang, and Q.~Tian, ``Rethinking rotated object detection with gaussian wasserstein distance loss,'' in \emph{International Conference on Machine Learning}.\hskip 1em plus 0.5em minus 0.4em\relax PMLR, 2021, pp. 11\,830--11\,841.

\bibitem{KLD}
X.~Yang, X.~Yang, J.~Yang, M.~Qi, W.~Wang, Q.~Tian, and J.~Yan, ``\BIBforeignlanguage{en-US}{Learning high-precision bounding box for rotated object detection via kullback-leibler divergence},'' \emph{\BIBforeignlanguage{en-US}{Neural Information Processing Systems,Neural Information Processing Systems}}, Dec 2021.

\bibitem{kfiou_iclr2022}
X.~Yang, Y.~Zhou, G.~Zhang, J.~Yang, W.~Wang, J.~Yan, X.~Zhang, and Q.~Tian, ``The kfiou loss for rotated object detection,'' in \emph{International Conference on Learning Representations}, 2022.

\bibitem{cfa_cvpr2021}
Z.~Guo, C.~Liu, X.~Zhang, J.~Jiao, X.~Ji, and Q.~Ye, ``Beyond bounding-box: Convex-hull feature adaptation for oriented and densely packed object detection,'' in \emph{Proceedings of the IEEE/CVF Conference on Computer Vision and Pattern Recognition}, 2021, pp. 8792--8801.

\bibitem{li2022orientedreppoints}
W.~Li, Y.~Chen, K.~Hu, and J.~Zhu, ``Oriented reppoints for aerial object detection,'' in \emph{Proceedings of the IEEE/CVF Conference on Computer Vision and Pattern Recognition}, 2022, pp. 1829--1838.

\bibitem{point_axes_eccv2024}
Z.~Zhao, Q.~Xue, Y.~He, Y.~Bai, X.~Wei, and Y.~Gong, ``Projecting points to axes: Oriented object detection via point-axis representation,'' in \emph{European Conference on Computer Vision}.\hskip 1em plus 0.5em minus 0.4em\relax Springer, 2024, pp. 161--179.

\bibitem{zhang2024orienteddino}
M.~Zhang, H.~Qiu, L.~Wang, H.~Cheng, T.~Zhao, and H.~Li, ``Oriented-dino: Angle decoupling prediction and consistency optimizing for oriented detection transformer,'' \emph{IEEE Transactions on Geoscience and Remote Sensing}, 2024.

\bibitem{li2022dndetr}
F.~Li, H.~Zhang, S.~Liu, J.~Guo, L.~M. Ni, and L.~Zhang, ``Dn-detr: Accelerate detr training by introducing query denoising,'' in \emph{Proceedings of the IEEE/CVF Conference on Computer Vision and Pattern Recognition}, 2022, pp. 13\,619--13\,627.

\bibitem{hu2023dacdetr}
Z.~Hu, Y.~Sun, J.~Wang, and Y.~Yang, ``Dac-detr: Divide the attention layers and conquer,'' \emph{Advances in Neural Information Processing Systems}, vol.~36, pp. 75\,189--75\,200, 2023.

\bibitem{zhao2024msdetr}
C.~Zhao, Y.~Sun, W.~Wang, Q.~Chen, E.~Ding, Y.~Yang, and J.~Wang, ``Ms-detr: Efficient detr training with mixed supervision,'' in \emph{Proceedings of the IEEE/CVF Conference on Computer Vision and Pattern Recognition}, 2024, pp. 17\,027--17\,036.

\bibitem{hu2023emo2detr}
Z.~Hu, K.~Gao, X.~Zhang, J.~Wang, H.~Wang, Z.~Yang, C.~Li, and W.~Li, ``Emo2-detr: Efficient-matching oriented object detection with transformers,'' \emph{IEEE Transactions on Geoscience and Remote Sensing}, 2023.

\bibitem{lin2017focal}
T.-Y. Lin, P.~Goyal, R.~Girshick, K.~He, and P.~Doll{\'a}r, ``Focal loss for dense object detection,'' in \emph{Proceedings of the IEEE International Conference on Computer Vision}, 2017, pp. 2980--2988.

\bibitem{sasm_aaai2022}
L.~Hou, K.~Lu, J.~Xue, and Y.~Li, ``Shape-adaptive selection and measurement for oriented object detection,'' in \emph{Proceedings of the AAAI Conference on Artificial Intelligence}, vol.~36, no.~1, 2022, pp. 923--932.

\bibitem{psc_cvpr2023}
Y.~Yu and F.~Da, ``Phase-shifting coder: Predicting accurate orientation in oriented object detection,'' in \emph{Proceedings of the IEEE/CVF Conference on Computer Vision and Pattern Recognition}, 2023, pp. 13\,354--13\,363.

\bibitem{xu2020glidingvertex}
Y.~Xu, M.~Fu, Q.~Wang, Y.~Wang, K.~Chen, G.-S. Xia, and X.~Bai, ``Gliding vertex on the horizontal bounding box for multi-oriented object detection,'' \emph{IEEE Transactions on Pattern Analysis and Machine Intelligence}, vol.~43, no.~4, pp. 1452--1459, 2020.

\bibitem{atss_cvpr2020}
S.~Zhang, C.~Chi, Y.~Yao, Z.~Lei, and S.~Z. Li, ``Bridging the gap between anchor-based and anchor-free detection via adaptive training sample selection,'' in \emph{Proceedings of the IEEE/CVF Conference on Computer Vision and Pattern Recognition}, 2020, pp. 9759--9768.

\bibitem{yang2021r3det}
X.~Yang, J.~Yan, Z.~Feng, and T.~He, ``R3det: Refined single-stage detector with feature refinement for rotating object,'' in \emph{Proceedings of the AAAI Conference on Artificial Intelligence}, vol.~35, no.~4, 2021, pp. 3163--3171.

\bibitem{han2021redet}
J.~Han, J.~Ding, N.~Xue, and G.-S. Xia, ``Redet: A rotation-equivariant detector for aerial object detection,'' in \emph{Proceedings of the IEEE/CVF Conference on Computer Vision and Pattern Recognition}, 2021, pp. 2786--2795.

\bibitem{zhao2025rediffdet}
J.~Zhao, Z.~Ding, Y.~Zhou, H.~Zhu, W.-L. Du, and R.~Yao, ``Rediffdet: Rotation-equivariant diffusion model for oriented object detection,'' in \emph{Proceedings of the Computer Vision and Pattern Recognition Conference}, 2025, pp. 24\,429--24\,439.

\bibitem{11195847}
P.~Gao, X.~Lu, K.~Li, G.~Cheng, and X.~You, ``Multi-scale oriented object detection with focus error ellipse loss,'' \emph{IEEE Transactions on Geoscience and Remote Sensing}, pp. 1--1, 2025.

\bibitem{arc_cvpr2023}
Y.~Pu, Y.~Wang, Z.~Xia, Y.~Han, Y.~Wang, W.~Gan, Z.~Wang, S.~Song, and G.~Huang, ``Adaptive rotated convolution for rotated object detection,'' in \emph{Proceedings of the IEEE/CVF International Conference on Computer Vision}, 2023.

\bibitem{zhou2022mmrotate}
Y.~Zhou, X.~Yang, G.~Zhang, J.~Wang, Y.~Liu, L.~Hou, X.~Jiang, X.~Liu, J.~Yan, C.~Lyu, W.~Zhang, and K.~Chen, ``Mmrotate: A rotated object detection benchmark using pytorch,'' in \emph{Proceedings of the 30th ACM International Conference on Multimedia}, 2022, p. 7331–7334.

\bibitem{he2017maskrcnn}
K.~He, G.~Gkioxari, P.~Doll{\'a}r, and R.~Girshick, ``Mask r-cnn,'' in \emph{Proceedings of the IEEE International Conference on Computer Vision}, 2017, pp. 2961--2969.

\bibitem{chen2019hybrid}
K.~Chen, J.~Pang, J.~Wang, Y.~Xiong, X.~Li, S.~Sun, W.~Feng, Z.~Liu, J.~Shi, W.~Ouyang \emph{et~al.}, ``Hybrid task cascade for instance segmentation,'' in \emph{Proceedings of the IEEE/CVF Conference on Computer Vision and Pattern Recognition}, 2019, pp. 4974--4983.

\bibitem{liu2021swintransformer}
Z.~Liu, Y.~Lin, Y.~Cao, H.~Hu, Y.~Wei, Z.~Zhang, S.~Lin, and B.~Guo, ``Swin transformer: Hierarchical vision transformer using shifted windows,'' in \emph{Proceedings of the IEEE/CVF International Conference on Computer Vision}, 2021, pp. 10\,012--10\,022.

\end{thebibliography}
\newpage
\vfill
\end{document}